\documentclass{article} 
\PassOptionsToPackage{table}{xcolor} 
\IfFileExists{iclr2027_conference.sty}
  {\usepackage{iclr2027_conference,times}}
  {\usepackage{natbib}%
   \newcommand{\And}{\and}%
   }
\usepackage{amsthm}

\newtheorem{proposition}{Proposition}

\newtheorem{definition}{Definition}
\newtheorem{assumption}{Assumption}
\newtheorem{remark}{Remark}
\IfFileExists{math_commands.tex}{

\usepackage{amsmath,amsfonts,bm}

\def\eqref#1{equation~\ref{#1}}

\def\1{\bm{1}}

\DeclareMathAlphabet{\mathsfit}{\encodingdefault}{\sfdefault}{m}{sl}
\SetMathAlphabet{\mathsfit}{bold}{\encodingdefault}{\sfdefault}{bx}{n}

}{}
\usepackage{amsmath}
\usepackage{amssymb}
\usepackage{booktabs}
\renewcommand{\arraystretch}{0.90}
\usepackage{hyperref}
\usepackage{url}
\usepackage{enumitem}
\usepackage{graphicx}
\usepackage{float}
\usepackage{multirow}
\usepackage{booktabs}
\usepackage{xcolor}
\usepackage{array}
\newcolumntype{L}[1]{>{\raggedright\arraybackslash}p{#1}}

\definecolor{cfbest}{HTML}{DCE9F6}   
\definecolor{cfbase}{HTML}{EFEFEF}   
\definecolor{cfrow}{HTML}{F6F6F6}    
\definecolor{cfhead}{HTML}{E9EEF3}   
\definecolor{cfgain}{HTML}{16702F}   
\definecolor{cfloss}{HTML}{A81E12}   
\definecolor{cfarrow}{HTML}{808080}  
\definecolor{cfref}{HTML}{6B6B6B}    
\definecolor{cftbd}{HTML}{B3651B}    
\definecolor{cflink}{HTML}{1A4E8A}
\definecolor{cfcite}{HTML}{2E6B3E}
\definecolor{cfurl}{HTML}{8A3A1A}

\newcommand{\metup}{\textcolor{cfarrow}{\raisebox{0.15ex}{$\scriptstyle\uparrow$}}}
\newcommand{\metdn}{\textcolor{cfarrow}{\raisebox{0.15ex}{$\scriptstyle\downarrow$}}}

\newcommand{\best}[1]{\cellcolor{cfbest}#1}
\newcommand{\basecell}[1]{#1}
\newcommand{\gain}[1]{\textcolor{cfgain}{#1}}
\newcommand{\loss}[1]{\textcolor{cfloss}{#1}}

\newcommand{\methodname}{\textsc{Multi-View Edit}}

\newcommand{\safeincludegraphics}[2][]{%
  \IfFileExists{#2}{\includegraphics[#1]{#2}}{%
    \fbox{\parbox[c][0.20\textheight][c]{0.92\linewidth}{%
      \centering Missing figure file:\\\texttt{\detokenize{#2}}}}}}

\title{Factual Access States in Diffusion Language Models: From Mechanism to Knowledge Editing and Retrieval}
\author{%
Yuchen Guo\thanks{This work was done during internship at LIGHTSPEED.} \\
The University of Tokyo \\
\texttt{guo-yuchen027@g.ecc.u-tokyo.ac.jp}
\And
Zhuoyuan Mao \\
LIGHTSPEED \\
\texttt{zhuoyuanmao@global.tencent.com}
\And
Yiwen Guo\thanks{Corresponding author.} \\
Independent Researcher \\
\texttt{guoyiwen89@gmail.com}
}

\hypersetup{pdfauthor={Yuchen Guo, Zhuoyuan Mao, Yiwen Guo},
  pdftitle={Factual Access States in Diffusion Language Models: From Mechanism to Knowledge Editing and Retrieval}}

\iclrfinalcopy 
\begin{document}

\maketitle
\lhead{Preprint}
\vspace{-0.12in}
\begin{center}\small\textbf{Code:} \url{https://github.com/Guoyuchen1/dlm-factual-access}\end{center}

\begin{abstract}
Factual recall is a lookup: some representation inside the network serves as
the key, and what that key is made of decides which queries can retrieve the
fact, what an edit must cover, and what a retriever can read. We call it the \textbf{factual access state}. In autoregressive (AR)
models the subject position enriches attributes largely independently of the
relation, and relation-specific selection happens later, at the readout
position. In masked diffusion language models (DLMs) we find the state is
assembled differently, and much earlier: in the first attention layer the
wording of the relation is routed into the position of whichever entity is
visible and the early feed-forward layers look the fact up on this fused representation.
The state therefore varies with the relation's surface form and with the query
direction, so the queries about one fact occupy a region of state space rather
than a point, and an operation on the fact succeeds only for queries that
enter it. We treat the state as an interface to a DLM's factual knowledge, and put it to two uses. \textbf{Writing} to the state gives a knowledge editor: from one
declarative statement, \methodname{} fits one shared value per direction
across that region and installs it in the early-MLP block, preserving
source-prompt reliability while substantially improving paraphrase and reverse
question answering, where AR editors gain near zero. \textbf{Reading} it
suggests a complementary use: the state at each entity position can serve as a
relation-specific retrieval vector with no training, returning both pieces of
evidence for two-entity queries on which pooled, prompted and lexical
retrievers almost never return both.
\end{abstract}


\section{Introduction}
\label{sec:intro}

\begin{figure*}[t]
  \centering
  \safeincludegraphics[width=\textwidth]{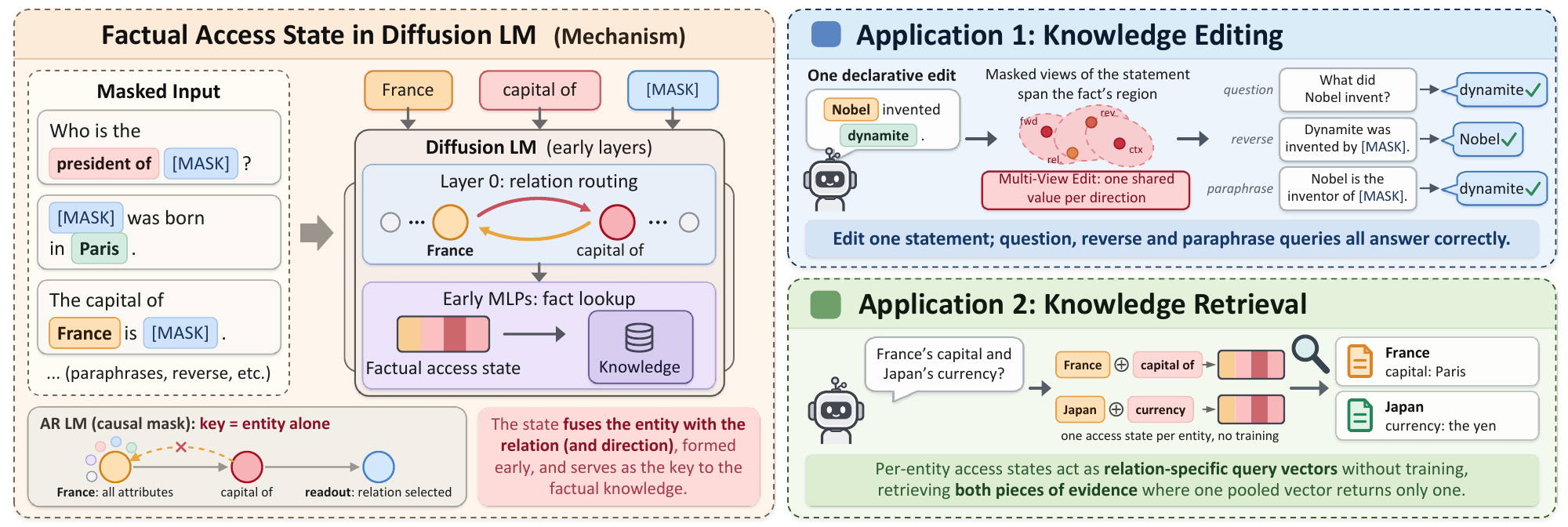}
  \caption{\textbf{Factual access states.} A masked DLM fuses an entity with
  the queried relation early; the resulting state supports knowledge editing
  and evidence retrieval.}
  \label{fig:factual-access-overview}
\end{figure*}

A language model recalls a fact by using an internal representation as a lookup
key. We call it the \emph{factual access state}: its composition determines
which queries reach a fact, which states an edit must cover, and what a
retriever can read. Knowledge editing changes the value associated with a
key~\citep{meng2022rome,meng2023memit}; evidence retrieval succeeds when query
and evidence representations align. Figure~\ref{fig:factual-access-overview}
previews the mechanism, its contrast with an AR key, and its two uses.

In autoregressive (AR) models, causal masking separates entity access from
relation selection. Subject positions accumulate attributes without seeing
relation tokens to their right; relation-specific selection occurs later at
the final position~\citep{geva2023dissecting,meng2022rome}. Locate-then-edit
methods therefore update an entity key at the last subject token and recover
it from similar prefixes. This is also their limit: the edit is reachable
from near-copies of the prefix, but not reliably from paraphrases or reverse
questions~\citep{berglund2023reversal,kitouni2024factorization,yang2025mirage}.

Masked diffusion language models (DLMs) break this premise. They predict any
masked subset with bidirectional attention~\citep{sahoo2024mdlm,nie2025llada,ye2025dream},
so a subject can see relation tokens to its right and an object can become a
key in its own direction. Recent work transfers AR edit sites to DLMs and
makes edits robust to diffusion decoding~\citep{park2026editing,yao2026timerome}.
Such ports retain edit-prompt reliability, yet generalise poorly to
paraphrases and leave reverse retrieval unchanged. If the key were still the
entity alone, these failures would be hard to explain. We therefore ask what
representation actually drives lookup (App.~\ref{app:related-work}).

Mask-native causal tracing (\S\ref{sec:mechanism}) shows that layer-$0$
attention routes relation wording into the visible entity; those edges carry
$84\%$ of the layer's causal effect. MLPs $1$--$8$ then retrieve the fact from
the fused state. The pipeline replicates in a second DLM and with subject and
object roles swapped. The state therefore depends on the entity, the relation's
surface form, and query direction: queries about one fact occupy a region of
state space rather than a point. An operation on the fact succeeds only when
a query's state enters that region.

A single-view editor fits one point; \methodname{} (\S\ref{sec:method}) instead
builds masked forward, reverse, relation, and contextual views, fitting one
shared value per direction in the early MLPs. Source-prompt reliability remains at least
$97.6\%$ on CounterFact and BAKE. On BAKE, reverse question answering rises
from $22.4\%$ to $63.2\%$ on LLaDA-8B and from $20.3\%$ to $83.6\%$ on
Dream-7B, while published AR editors gain under three points. On QAEdit,
$32.5\%$ and $44.6\%$ of previously wrong free-form answers become at least
partially correct; unrelated-query activation is unchanged.

To read the state, we use each entity position as a relation-specific retrieval
vector (\S\ref{sec:retrieval}). Without training or question decomposition,
these vectors retrieve both pieces of evidence for two-entity queries on which
a pooled vector usually returns one. Removing the layer-$0$ relation messages
eliminates the advantage.

Our contributions are:

\begin{itemize}[leftmargin=1.2em,itemsep=1pt,topsep=2pt,parsep=0pt]
  \item \textbf{Mechanism of factual access states.} Causal tracing and
  attention surgery locate the relation-fused entity state, with replication
  across two DLMs and both mask directions.
  \item \textbf{Writing to the state: multi-view editing.} A shared value per
  direction covers the statement's query views; matched-budget ablations and
  key-space analysis account for the gains.
  \item \textbf{Reading the state: evidence retrieval.} Per-entity states
  retrieve both facts in two-entity queries, with early readout and causal
  support from relation-routing interventions.
\end{itemize}

\section{How a Masked DLM Accesses a Fact}
\label{sec:mechanism}
\newcommand{\Mask}{\texttt{[MASK]}}

We first locate factual lookup in LLaDA-8B (\S\ref{sec:footprint}) and show
how layer-$0$ attention assembles the state read by the early MLPs
(\S\ref{sec:l0-surgery}). The findings replicate on Dream-7B
(App.~\ref{app:dream-repro}) and when the object is visible instead of the
subject (App.~\ref{app:mask-subject}).

\subsection{Preliminaries}
\label{sec:prelim}
\label{sec:dlm}

\paragraph{Masked diffusion language models.}
A masked DLM $p_\theta$ reconstructs randomly hidden tokens. Given $x$ and a
mask rate $t\!\sim\!\mathcal{U}(0,1)$, the corruption operator $q_t$ replaces
each token independently by \Mask{} with probability $t$, and the model is
trained to recover the hidden tokens from the visible ones with bidirectional
attention (the masked-diffusion ELBO; App.~\ref{app:theory-setup}). Two
consequences matter here. Where an AR model exposes only $p(o\mid s,r)$, a
masked DLM exposes a whole family of conditionals about the same fact, any
subset conditioned on and any subset predicted; this is what
\S\ref{sec:method} exploits. And because the model is queried under $q_t$ at
both training and decoding time, the states it reaches at test time come
from partially masked inputs, which is why the tracing below corrupts with
\Mask{} rather than with noise.

\paragraph{Mask-based causal tracing.}
\label{sec:tracing}
Causal tracing corrupts the input until the answer is lost, then restores
internal states one at a time to see which bring it back. We adapt ROME's
formulation~\citep{meng2022rome} in two places: next-token prediction becomes
masked-position prediction, and Gaussian noise becomes the native \Mask{}
corruption. For each CounterFact~\citep{meng2022rome} triple $(s,r,o)$ the
\emph{clean} input $x^{\mathrm{clean}}$ masks the object span, so the task is
exactly what the model was trained on, and the \emph{corrupted} input
$x^{\mathrm{corr}}$ masks both subject and object, removing the cue.
Restoring a hidden state $h_\ell^{(i)}$ to its clean value during the
corrupted pass defines the \emph{indirect effect}
\begin{equation}
\label{eq:ie}
\mathrm{IE}(\ell, i)
\;=\;
\frac{\,
   p_{\mathrm{rest}}\!\bigl(o \mid x^{\mathrm{corr}},\;
      h_\ell^{(i)}\!\leftarrow h_\ell^{\mathrm{clean},(i)}\bigr)
   - p\!\bigl(o \mid x^{\mathrm{corr}}\bigr)\,
}{\,
   p\!\bigl(o \mid x^{\mathrm{clean}}\bigr)
   - p\!\bigl(o \mid x^{\mathrm{corr}}\bigr)\,
},
\end{equation}
where $\mathrm{IE}{=}1$ means that state alone recovers the clean answer and
$0$ means it contributes nothing.

\subsection{Where the fact is looked up}
\label{sec:footprint}
\label{sec:bimodal}
\label{sec:stacking}
\label{sec:early-block}

\begin{figure}[htbp]
  \centering
  \safeincludegraphics[width=0.98\linewidth]{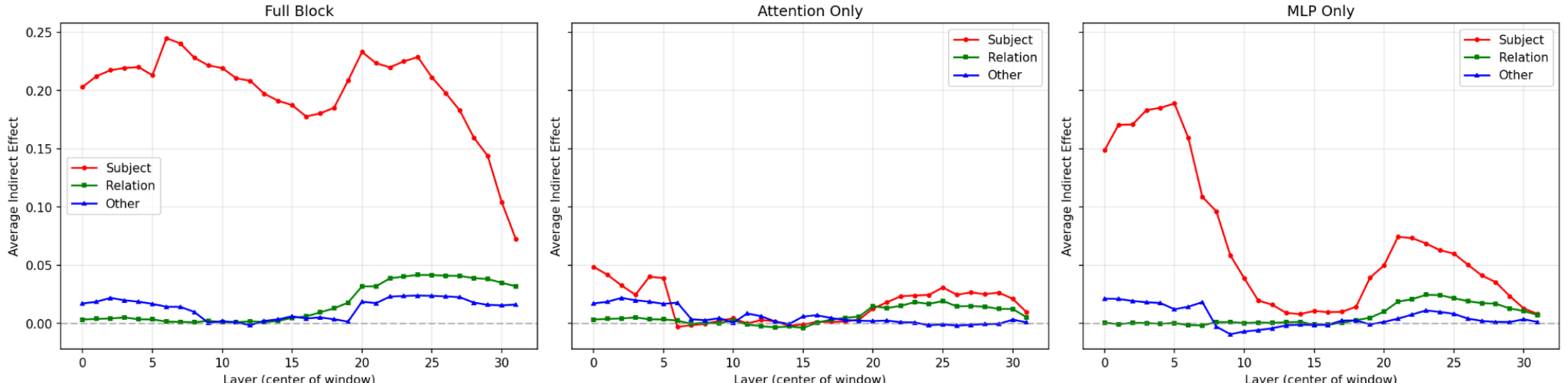}
  \caption{\textbf{Window-tracing on a masked DLM (LLaDA-8B).}
  Sliding-window causal tracing with $w{=}10$, showing the indirect effect
  across layers at subject, relation, and other token positions. Left: full
  block; middle: attention only; right: FF/MLP only. The subject signal is
  bimodal and MLP-dominated.}
  \label{fig:trace-window}
\end{figure}

Restoring a sliding window of $w{=}10$ adjacent layers
(Fig.~\ref{fig:trace-window}) concentrates the effect almost exclusively on the
\emph{subject} tokens. The profile is \textbf{bimodal}, with an early peak at
layers $0$--$8$ and a late peak at $18$--$25$, both MLP-dominated. This mirrors
a \emph{retrieve-then-decode} view.

\begin{figure}[htbp]
  \centering
  \safeincludegraphics[width=0.98\linewidth]{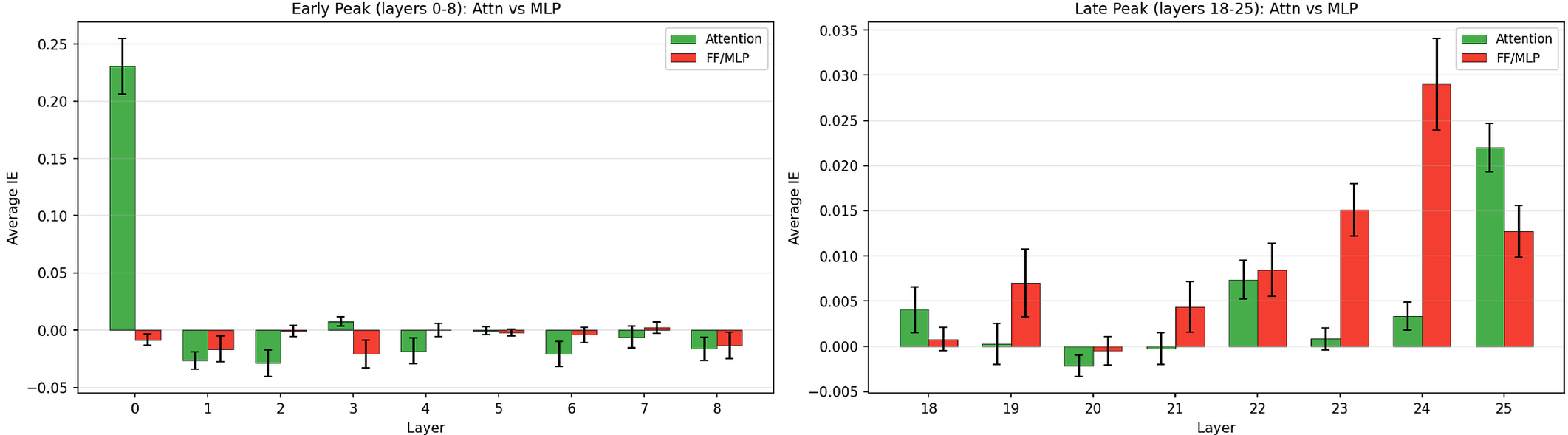}
  \caption{\textbf{Single-layer causal effect.}
  Per-layer IE inside the early ($0$--$8$) and late ($18$--$25$) peaks,
  split by attention vs.\ MLP. Only layer-$0$ attention stands out; no
  single MLP layer exceeds $\sim\!0.03$.}
  \label{fig:single-layer}
\end{figure}

The second peak is an artefact of window stacking. When a \emph{single} layer
is ablated at a time (Fig.~\ref{fig:single-layer}), only \textbf{layer-0
attention} produces a large effect ($\mathrm{IE}\!\approx\!0.23$), while every
individual MLP layer contributes $\lesssim\!0.03$. The late peak is a sum of
many small redundant contributions, not a second retrieval site.

What survives is an early block that is jointly, but only jointly, sufficient.
Restoring $\ell{=}0{-}8$ reaches $\mathrm{IE}{=}0.936$, essentially the
whole-network ceiling. Layer-$0$ attention alone ($0.238$), MLPs $1$--$8$ alone
($0.781$) and layer-$0$ MLP alone ($-0.015$) each fall short, but layer-$0$
attention combined with MLPs $1$--$8$ recovers $0.943$
(Fig.~\ref{fig:early-decomp}). This is consistent with early-MLP retrieval
being conditioned on what layer-$0$ attention writes into the subject
positions, and the block is the site that \S\ref{sec:method} writes into.
App.~\ref{app:causal-extra} adds two further analyses: why the late peak is a
serial conduit rather than a second retrieval site, and how the causal locus
migrates from subject tokens to the \Mask{} readout position.

\subsection{Layer 0 assembles the access state}
\label{sec:l0-surgery}

The trace says layer $0$ matters but not what it computes. We therefore run
five \emph{attention-surgery} ablations: in each, subject tokens may attend
along only one class of edges and all others are zeroed
(Table~\ref{tab:l0-surgery}).

\begin{table}[htbp]
\centering
\footnotesize
\caption{Layer-$0$ attention surgery on LLaDA-8B. Each row keeps only the
listed class of edges out of the subject tokens; \% is normalised by the full
layer-$0$ attention IE of $0.238$.}
\label{tab:l0-surgery}
\begin{tabular}{lcc}
\toprule
Layer-$0$ restriction on subject tokens
    & IE (mean\,$\pm$\,sem)~\metup
    & \% of full L0 attn~\metup \\
\midrule
sink removed (\texttt{sink\_off})           & $+0.187 \pm 0.047$ & $79\%$ \\
\best{subject $\to$ \textbf{relation} only} & \best{$\mathbf{+0.199 \pm 0.044}$} & \best{$\mathbf{84\%}$} \\
subject $\to$ subject-internal only         & $+0.049 \pm 0.035$ & $21\%$ \\
subject $\to$ self only                     & $+0.008 \pm 0.010$ & $3\%$  \\
subject $\to$ \Mask{} only                  & $-0.008 \pm 0.010$ & $0\%$  \\
\bottomrule
\end{tabular}
\end{table}

Letting subject tokens read only from the relation template recovers $84\%$ of
the effect, while subject-internal binding, self-attention and the sink are
each dispensable. Layer-$0$ attention thus \emph{merges the relation wording
into the subject residual stream}, producing the relation-sensitive
representation that the layer-$1$--$8$ MLPs read. We give this representation
a name.

\begin{definition}[Factual access state]
\label{def:access-state}
For a query in which entity $e$ is visible and the relation is worded as $r$,
the \emph{factual access state} $a(e,r,d)$ is the residual-stream
representation at the position of $e$ in the early block, after layer-$0$
attention has routed the wording of $r$ into it; $d\in\{\mathrm{fwd},
\mathrm{rev}\}$ records whether $e$ is the subject or the object of the fact.
The early MLPs read it as their input key, $k^{(\ell)}(\cdot,p_e)$ being its
image at layer $\ell$; \S\ref{sec:method} writes at $k^{(7)}$ and
\S\ref{sec:retrieval} reads $a$ at the output of block~$0$ or at a deeper layer. Because layer $0$
routes the relation into $e$, the state is a function of all three arguments,
\begin{equation}
  a \;=\; f\big(e,\;\text{surface form of } r,\;d\big),
  \label{eq:binding}
\end{equation}
not of the entity alone.
\end{definition}

The routing is bidirectional, and so is the state. When the object is kept
visible and the subject is hidden instead, the same pipeline runs with the
roles swapped: the object position now carries the early-block hidden and
MLP effect, its layer-$0$ attention spikes in the same way, and relation and
prefix positions stay flat (App.~\ref{app:mask-subject}). The state is
therefore formed at whichever entity is visible, with the relation routed
into it from either side. Under a causal mask none of this is available: a
subject position cannot see a relation to its right, so an autoregressive
model keys a fact on the subject and the prefix alone, forms no state on the
object side, and cannot express a reverse association natively.

\paragraph{Take-aways.}
The factual access state has three properties that guide the rest of the
paper. First, it depends on the relation's surface form, so queries
about the same fact occupy a region of state space rather than a point.
Second, it forms at whichever entity is visible, enabling both forward
and reverse access. Third, it is available from the first block and
refined by deeper layers, making early readout possible.
These properties motivate the multi-view editing method in
\S\ref{sec:method} and suggest a complementary use for evidence
retrieval in \S\ref{sec:retrieval}.
The state is also present on fully observed inputs, as confirmed in
\S\ref{sec:retrieval} and App.~\ref{app:retrieval-causal}.

\section{Writing to the State: Multi-View Editing from a Single Statement}
\label{sec:method}
\newcommand{\Vobj}{\mathcal{V}_{\mathrm{obj}}}
\newcommand{\Vrel}{\mathcal{V}_{\mathrm{rel}}}
\newcommand{\Vctx}{\mathcal{V}_{\mathrm{ctx}}}
\newcommand{\Wdown}{W_{\mathrm{down}}}

\methodname{} follows \S\ref{sec:mechanism} on both counts: it leaves
attention untouched, since the routing is shared by every query about the
fact, and rewrites the early MLPs using a family of masked views covering the
access states the same statement induces at test time.

\subsection{Method}
\label{sec:method:method}

\subsubsection{Editing task and the states an edit must cover}
\label{sec:method:principle}

A fact $(s,r,o)$ is given as a statement $x$ with spans
$\mathcal{S}_s,\mathcal{S}_r,\mathcal{S}_o$ and a new object $o^{*}$; the goal
is to modify $\theta$ so the model behaves as if $(s,r,o^{*})$ held, leaving
unrelated behaviour intact. Following \S\ref{sec:mechanism}, we edit the
down-projections $\Wdown^{(\ell)}$ of the early block, $\ell \in \mathcal{L} =
\{\ell_1,\dots,\ell_L\} \subset [1,8]$. For input $u$ at position $p$,
$k^{(\ell)}(u,p) \in \mathbb{R}^{d_{\mathrm{mlp}}}$ denotes the input to
$\Wdown^{(\ell)}$, that is, the representation retrieval reads: the \emph{key}.
Its output $z^{(\ell)}(u,p) = \Wdown^{(\ell)}k^{(\ell)}(u,p)$ is the
\emph{value}. An edit installs one key--value association.

By Definition~\ref{def:access-state}, the key at an entity position is the
access state of that query as the layer-$\ell$ MLP reads it, so it depends on
the relation's surface form and on the query direction, \eqref{eq:binding}:
an edit fitted to the user's sentence covers one point in key space, and
rephrased or reverse queries activate it weakly. Our design principle is
therefore:
\begin{quote}
\emph{Fit the edit to the average of the keys the fact will be queried from,
read from either side of it, rather than to the key of the one sentence it
happens to be written in.}
\end{quote}
App.~\ref{app:theory} formalises this principle as expected activation over
the fact's forward and reverse query families and analyses why the view family
approximates the desired keys.

\subsubsection{View family construction}
\label{sec:method:views}

We construct views with the model's own corruption operator. Writing
$\mathcal{M}(u;\mathcal{P})$ for the input with tokens at $\mathcal{P}$
replaced by \Mask{}, the \emph{forward} family hides the object and reads the
key at a subject position,
\begin{align}
  \Vobj &: \;\; \mathcal{M}(x;\mathcal{S}_o),
  &&\text{supervise } o^{*} \text{ at the masked object span,} \label{eq:vobj}\\
  \Vrel(\rho) &: \;\; \mathcal{M}\big(x;\mathcal{S}_o \cup \mathcal{R}_\rho\big),
  \, \mathcal{R}_\rho \subset \mathcal{S}_r,\;
  |\mathcal{R}_\rho| = \lceil \rho\,|\mathcal{S}_r| \rceil,
  &&\text{relation partly masked,} \label{eq:vrel}\\
  \Vctx &: \;\; \mathcal{M}\big(c_{\mathrm{pre}} \oplus x \oplus c_{\mathrm{post}};\,\mathcal{S}_o\big),
  &&\text{prefix/suffix context templates,} \label{eq:vctx}
\end{align}
and the \emph{reverse} family swaps the roles, replacing the object span with
the new object and hiding the subject, so the key is read at $o^{*}$:
\begin{equation}
  \mathcal{V}^{\mathrm{rev}} : \;\;
  \mathcal{M}\big(x[\mathcal{S}_o \!\leftarrow\! o^{*}];\;
  \mathcal{S}_s \cup \mathcal{R}_\rho \big),
  \qquad\text{supervise } s \text{ at the masked subject span,}
  \label{eq:revview}
\end{equation}
For $\Vrel$ and $\mathcal{V}^{\mathrm{rev}}$ we draw $n_\rho$ occlusion
patterns per ratio $\rho \in \mathcal{P}_\rho$ ($\mathcal{P}_\rho =
\{0.4,0.7\}$, $n_\rho = 3$). Whichever span carries the key is never masked.
ROME averages the key over a set of prefix templates so that it does not depend
on the text around the prompt; $\Vctx$ does the same. Because a DLM also
conditions on what follows the prediction slot, $c_{\mathrm{post}}$ may be
non-empty, and we add suffix templates alongside the prefix ones
(App.~\ref{app:hyperparams}).

Two axes organise these views, both instantiated by the same $q_t$. The
\emph{direction} axis decides how many key--value directions are installed:
$\mathcal{V}^{\mathrm{rev}}$ adds the object-side one and, by
\eqref{eq:revview}, carries relation occlusions on that side too. The
\emph{masking} axis decides how much of each installed direction a
reformulated query reaches: $\Vobj$ fixes the association, $\Vctx$
decorrelates the key from the surrounding discourse, and $\Vrel$ decorrelates
it from the relation's surface form, which by \eqref{eq:binding} is what
layer-$0$ merges in. Together they form a Monte-Carlo sample, restricted to
this fact, of the conditionals the model is trained on under $q_t$. Setting
$\mathcal{P}_\rho = \emptyset$
leaves $\Vobj \cup \Vctx$. This is the direct port of a locate-then-edit
editor, which already averages keys over context templates, and it is the
\textsc{Single-View} baseline of Table~\ref{tab:cf}; those two rows therefore
isolate $\Vrel$. Table~\ref{tab:ablation-context} ablates $\Vctx$ separately.

\subsubsection{One shared value per direction}
\label{sec:method:objective}

Let $\mathcal{V} = \Vobj \cup \Vrel \cup \Vctx$ and $\ell^{*} = \max
\mathcal{L}$. Writing $p_{\theta[\ell^{*}\leftarrow z]}$ for the model with
$\Wdown^{(\ell^{*})}$'s output at the key position replaced by $z$, we solve
\begin{equation}
  z^{*}_{\mathrm{fwd}}
  \;=\;
  \arg\min_{z}\;
  \frac{1}{|\mathcal{V}|}\sum_{v \in \mathcal{V}} w_v
  \Big[-\log p_{\theta[\ell^{*}\leftarrow z]}\big(o^{*} \,\big|\, \tilde{x}_v\big)\Big]
  \;+\;
  v_{\mathrm{wd}}\,\frac{\lVert z - z_0 \rVert^{2}}{\lVert z_0 \rVert^{2}},
  \qquad
  \lVert z - z_0 \rVert \le \kappa \lVert z_0 \rVert ,
  \label{eq:zstar}
\end{equation}
with $z_0$ the unedited value, $w_v$ per-view weights (uniform here, since all
views are drawn at comparable mask rates), $v_{\mathrm{wd}}$ a relative norm
penalty and $\kappa$ a clamp. Adam optimises $z$ alone; model parameters
stay frozen. $z^{*}_{\mathrm{rev}}$ is obtained the same way over
$\mathcal{V}^{\mathrm{rev}}$, averaging the loss over masked subject positions
instead of the object span.

One value is shared across all views of a direction, making \eqref{eq:zstar} an
\emph{intersection} rather than a union: one $(k,z)$ pair per view would
install $|\mathcal{V}|$ separate associations, each reachable only from its own
point of key space. Instead we ask for a single value that is correct when read
out from every key the fact induces. Hard views therefore reshape $z^{*}$ as
constraints rather than becoming separate edits, and the installed rank does
not grow with $|\mathcal{V}|$.

\subsubsection{Keys and the weight update}
\label{sec:method:update}

With $p_v$ the key position in view $v$ (a subject position forward, the
chosen token of $o^{*}$ in reverse), the key written along is the average
over the same views used in \eqref{eq:zstar},
\begin{equation}
  k_{\mathrm{avg}}
  \;=\;
  \frac{1}{|\mathcal{V}|}\sum_{v\in\mathcal{V}} k^{(\ell)}\big(\tilde{x}_v, p_v\big),
  \label{eq:kavg}
\end{equation}
so that the direction itself is averaged over occlusions and contexts;
$k^{\mathrm{rev}}_{\mathrm{avg}}$ follows identically over
$\mathcal{V}^{\mathrm{rev}}$. The update is applied along the whitened
direction $\hat{k} = C^{-1}k_{\mathrm{avg}} /
(k_{\mathrm{avg}}^{\top}C^{-1}k_{\mathrm{avg}})$, where $C$ is the second
moment of keys collected from randomly masked inputs under the same $q_t$ the
model traverses during decoding, so that the edit does not fire along
directions that are already high-variance across ordinary text
(App.~\ref{app:update-details}; ablated in App.~\ref{app:ablation}). With
$\ell_i$ the $i$-th edit layer and $m = L-i+1$ the remaining layers, the
residual is distributed across $\mathcal{L}$ as in MEMIT:
\begin{equation}
  \Wdown^{(\ell_i)} \;\leftarrow\;
  \Wdown^{(\ell_i)} \;+\;
  \frac{z^{*} - \Wdown^{(\ell_i)}k_{\mathrm{avg}}}{m}\;\hat{k}^{\top}.
  \label{eq:rank1}
\end{equation}
Forward and reverse pairs go into the same layers, so one statement installs
one forward and one or more reverse directions in one early-MLP block.
Attention is never modified: the problem lies in \emph{what the state is made
of}, which the view family addresses, not in the mechanism that assembles it.

\subsection{Results}
\label{sec:experiments}
\label{sec:setup}

We evaluate on CounterFact (source reliability, paraphrase transfer,
neighborhood specificity and locality preservation), BAKE (reverse QA and
judgment probes) and QAEdit (free-form answers to natural questions, scored
exact, partial or wrong), on LLaDA-8B and Dream-7B. ROME at layer $7$ is the
base editor, constrained fine-tuning (FT+L) the reproducible DLM baseline, and
matched single-view and single-direction variants the controlled comparisons.
Benchmark definitions, scoring, configuration and published autoregressive
references are in App.~\ref{app:editing-details}.

\subsubsection{CounterFact}
\label{sec:cf}

Table~\ref{tab:cf} isolates the relation-masked views against the matched
single-view editor. Multi-View keeps source reliability high and consistently
improves paraphrase transfer. On LLaDA it also improves neighborhood
specificity and locality preservation; on Dream, locality is comparable to
Single-View, indicating model-dependent spillover. Locality does not degrade
under the wider view family, which agrees with the neighborhood trigger
strength of \S\ref{sec:multiview-analysis}: it is unchanged between the two
constructions.

\begin{table}[htbp]
\centering
\footnotesize
\setlength{\tabcolsep}{3.5pt}
\caption{\textbf{CounterFact.} ROME-based rows use a last-subject-token
write; FT+L is a constrained fine-tuning baseline. \textsc{Single-View} uses
$\Vobj\cup\Vctx$; \textsc{Multi-View} adds $\Vrel$. Published AR references
are collected in App.~\ref{app:ar-refs}.}
\label{tab:cf}
\begin{tabular}{llcccc}
\toprule
\textbf{Model} & \textbf{Method}
  & Rel.~\metup & Para.~\metup & Neigh.~\metup & Loc-Pres.~\metup \\
\midrule
\multirow{3}{*}{Dream-7B}
  & FT+L
  & 99.67 & 86.47 & 18.95 & 31.33 \\
  & \textsc{Single-View}
  & 94.83 & 72.05 & \best{\textbf{74.48}} & \best{\textbf{89.91}} \\
  & \textsc{Multi-View}
  & \best{\textbf{99.88}}& \best{\textbf{99.24}} & 73.04 & 88.29 \\
\midrule
\multirow{3}{*}{LLaDA-8B}
  & FT+L
  & \best{\textbf{99.27}} & 51.94 & 47.28 & 65.04 \\
  & \textsc{Single-View}
  & 96.87 & 53.65 & 57.13 & 68.31 \\
  & \textsc{Multi-View}
  & 98.14 & \best{\textbf{67.92}} & \best{\textbf{73.73}} & \best{\textbf{89.92}} \\
\bottomrule
\end{tabular}
\vspace{2pt}
\end{table}

\subsubsection{BAKE: Reverse Access}
\label{sec:bake}

The no-edit row of Table~\ref{tab:bake} measures the model's pre-existing
preference for the target. \textsc{Single-Forward} edits only the source
direction; \textsc{Single-Binary} adds the reverse direction; and
\textsc{Multi-Binary} replaces each with the masked view family. BAKE uses a
different edit pool from CounterFact. The two axes separate cleanly.
Forward-only editing leaves Reverse-QA at the no-edit baseline: bidirectional
attention makes the reverse conditional expressible, but a forward update is
not reachable from it, so the reverse direction has to be installed
explicitly. Once it is, the masking axis governs how much of that direction a
reverse query reaches. It adds $21.2$ points on LLaDA and $10.6$ on Dream
($21.8$ at a matched view budget, Table~\ref{tab:ablation-budget-bake}) and
recovers the paraphrase accuracy that \textsc{Single-Binary} gives up.
Published autoregressive editors gain under three points on Reverse-QA on
GPT-J and LLaMA-2 (App.~\ref{app:ar-refs}).

\begin{table}[htbp]
\centering
\footnotesize
\setlength{\tabcolsep}{4pt}
\caption{\textbf{BAKE.} R-QA and R-Judge are reverse probes;
$\Delta$ is measured from each DLM's no-edit baseline. Single/Multi denotes one
view versus the masked family; Forward/Binary denotes whether the explicit
reverse group is installed. FT+L is a constrained fine-tuning baseline.
Published AR references are collected in App.~\ref{app:ar-refs}.}
\label{tab:bake}
\begin{tabular}{llccccc}
\toprule
\textbf{Model} & \textbf{Method}
  & Rel.~\metup & Para.~\metup & R-QA~\metup & R-Judge~\metup
  & $\boldsymbol{\Delta}$ \textbf{(R-QA / R-Judge)}~\metup \\
\midrule
\multirow{5}{*}{LLaDA-8B}
  & \basecell{\emph{no edit (baseline)}}
  & \basecell{--} & \basecell{--} & \basecell{22.35} & \basecell{30.59}
  & \basecell{$0.00\,/\,0.00$} \\
  & FT+L
  & 74.76 & 45.73 & 19.94 & 28.79
  & \loss{$-2.41$}\,/\,\loss{$-1.80$} \\
  & \textsc{Single-Forward}
  & 95.53 & 73.03 & 22.52 & 38.88
  & \gain{$+0.17$}\,/\,\gain{$+8.29$} \\
  & \textsc{Single-Binary}
  & 97.04 & 65.48 & 42.03 & 41.03
  & \gain{$+19.68$}\,/\,\gain{$+10.44$} \\
  & \textsc{Multi-Binary}
  & \best{\textbf{98.49}} & \best{\textbf{84.56}}
  & \best{\textbf{63.24}} & \best{\textbf{55.60}}
  & \gain{$\mathbf{+40.89}$}\,/\,\gain{$\mathbf{+25.01}$} \\
\midrule
\multirow{5}{*}{Dream-7B}
  & \basecell{\emph{no edit (baseline)}}
  & \basecell{--} & \basecell{--} & \basecell{20.26} & \basecell{18.15}
  & \basecell{$0.00\,/\,0.00$} \\
  & FT+L
  & 92.91 & 72.82 & 18.42 & 19.42
  & \loss{$-1.84$}\,/\,\gain{$+1.27$} \\
  & \textsc{Single-Forward}
  & 94.93 & 88.43 & 20.13 & 32.13
  & \loss{$-0.13$}\,/\,\gain{$+13.98$} \\
  & \textsc{Single-Binary}
  & 94.10 & 87.45 & 72.97 & 44.12
  & \gain{$+52.71$}\,/\,\gain{$+25.97$} \\
  & \textsc{Multi-Binary}
  & \best{\textbf{97.65}} & \best{\textbf{96.25}}
  & \best{\textbf{83.57}} & \best{\textbf{55.75}}
  & \gain{$\mathbf{+63.31}$}\,/\,\gain{$\mathbf{+37.60}$} \\
\bottomrule
\end{tabular}
\end{table}

\subsubsection{QAEdit: Open-Ended Generation}
\label{sec:qaedit}

QAEdit tests the statement-level setting directly: the edit is declarative,
whereas evaluation uses an open-ended question and free decoding. For each
model and method we evaluate the same $2{,}272$ examples that were answered
fully wrongly before editing (Table~\ref{tab:qaedit}). Multi-View gives the
strongest source-prompt recovery and question transfer on both models,
outperforming FT+L and Single-View on the principal success metrics. The gains
are larger on Dream, but free-generation transfer remains incomplete: wrong
answers are still the majority on the Gen.\ split (App.~\ref{app:limitations}).
The view family also composes with the decoding-state views of
\citet{park2026editing}: adding them raises the LLaDA-8B weighted score from
$0.250$ to $0.430$ on source prompts and from $0.187$ to $0.352$ on paraphrased
questions (App.~\ref{app:compose-parkjo}).

\begin{table}[htbp]
\centering
\footnotesize
\setlength{\tabcolsep}{4pt}
\caption{\textbf{QAEdit on pre-edit-wrong examples.}
Fraction of answers scored exact ($=1.0$) or at least partially
correct ($\ge 0.5$), and weighted score
$s=(n_{1.0}+0.5n_{0.5})/n$.
Rel.\ denotes source prompts and Gen.\ paraphrased questions.
Target-length results are in App.~\ref{app:qaedit-length}.}
\label{tab:qaedit}

\resizebox{\linewidth}{!}{%
\begin{tabular}{llcccccc}
\toprule
& & \multicolumn{3}{c}{\textbf{Rel.}}
    & \multicolumn{3}{c}{\textbf{Gen.}} \\
\cmidrule(lr){3-5}\cmidrule(lr){6-8}
\textbf{Model} & \textbf{Method}
& $=\!1.0$~\metup & $\ge\!0.5$~\metup & Weighted $s$~\metup
& $=\!1.0$~\metup & $\ge\!0.5$~\metup & Weighted $s$~\metup \\
\midrule
\multirow{3}{*}{LLaDA-8B}
& FT+L
& 15.0\% & 26.3\% & 0.206
& 11.9\% & 22.9\% & 0.174 \\
& Single-View
& 14.9\% & 26.2\% & 0.205
& 11.7\% & 23.3\% & 0.175 \\
& Multi-View
& \best{\textbf{17.6\%}} & \best{\textbf{32.5\%}} & \best{\textbf{0.250}}
& \best{\textbf{12.6\%}} & \best{\textbf{25.5\%}} & \best{\textbf{0.187}} \\
\midrule
\multirow{3}{*}{Dream-7B}
& FT+L
& 16.2\% & 32.4\% & 0.243
& 13.4\% & 26.8\% & 0.201 \\
& Single-View
& 20.1\% & 38.1\% & 0.291
& 16.5\% & 33.1\% & 0.248 \\
& Multi-View
& \best{\textbf{23.8\%}} & \best{\textbf{44.6\%}} & \best{\textbf{0.342}}
& \best{\textbf{20.2\%}} & \best{\textbf{39.3\%}} & \best{\textbf{0.297}} \\
\bottomrule
\end{tabular}%
}
\end{table}

\subsubsection{Edits Fire Where the State Is: A Trigger-Strength Analysis}
\label{sec:multiview-analysis}

\begin{figure}[htbp]
  \centering
  \safeincludegraphics[width=0.95\linewidth]{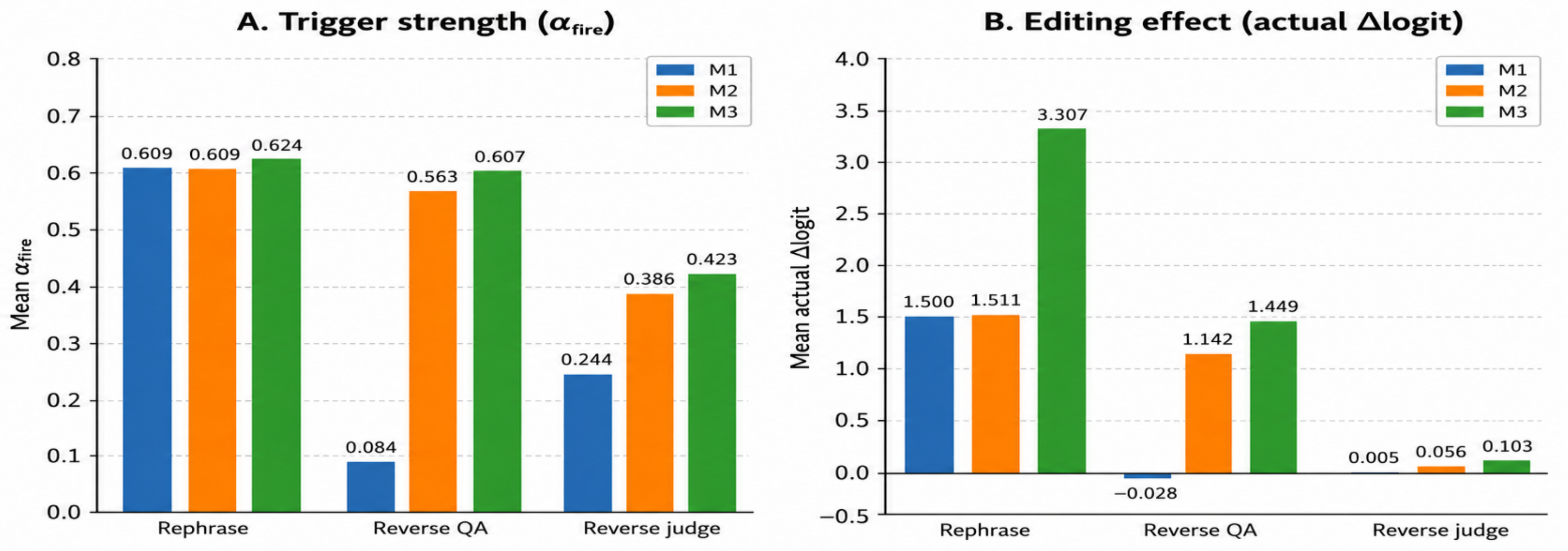}
  \caption{\textbf{Multi-View trigger and editing effect.} Left: mean
  trigger strength $\alpha_{\text{fire}}$; right: mean actual
  $\Delta\text{logit}$. M1 = forward single-view, M2 = forward $+$
  reverse single-view, M3 = full Multi-View (ours). LLaDA-8B. Query difficulty increases from Paraphrase (left) to Reverse-Judge
  (right).}
  \label{fig:multiview}
\end{figure}

We compare M1 (one forward key), M2 (forward and reverse keys), and M3 (both
directions expanded over the masked view family). For a query key $k$,
\eqref{eq:rank1} gives the \emph{trigger strength}
\begin{equation}
  \alpha(k) \;=\; \hat{k}^{\top}k
  \;=\;
  \frac{k_{\mathrm{avg}}^{\top}C^{-1}k}{k_{\mathrm{avg}}^{\top}C^{-1}k_{\mathrm{avg}}},
  \label{eq:alpha}
\end{equation}
the fraction of the installed update activated by the query. We report
$\alpha_{\text{fire}}$ at its most strongly activated position, alongside the
editing effect $\Delta\text{logit} = \log p_{\text{post}}(o_{\text{new}}) - \log
p_{\text{post}}(o_{\text{true}})$. On $n{=}1000$ BAKE edits with LLaDA-8B and
the ROME base editor, Fig.~\ref{fig:multiview} shows that reverse keys
strengthen reverse-query activation, while the full view family further
improves paraphrase and reverse probes: on Reverse-QA,
$\alpha_{\text{fire}}$ rises from $0.08$ with a forward key alone to $0.56$
once a reverse key is installed and to $0.61$ with the full family, while
paraphrase activation stays at $0.61$--$0.62$ throughout and its
$\Delta\text{logit}$ more than doubles. On CounterFact neighborhood prompts
about other subjects, $\alpha=0.03$ for both single-view and Multi-View edits,
well below the queried-fact values. The wider coverage preserves specificity.

\section{Reading the State: Evidence Retrieval}
\label{sec:retrieval}

\subsection{Method}
\label{sec:retrieval:method}

The access state is also an interface to external evidence. We read one
retrieval vector per query entity from its state $a(e,r,d)$, at the output of
block~$0$ (L0) or at a validation-selected deeper layer. For a token span
$\mathcal S$ in input $u$, write its layer-$\ell$ readout as
\begin{equation}
  v^{(\ell)}(u,\mathcal S)
  = \operatorname{pool}_{p\in\mathcal S} h_\ell^{(p)}(u).
  \label{eq:retrieval-readout}
\end{equation}
The pool selects the last entity token at L0 and averages the span at deeper
layers. For query $q$ with entity span $\mathcal S_{e_i}$ and evidence card
$x_j$ with subject span $\mathcal S_{s_j}$, we rank cards by
\begin{equation}
  \operatorname{score}_\ell(i,j)
  = \cos\!\left(v^{(\ell)}(q,\mathcal S_{e_i}),
                 v^{(\ell)}(x_j,\mathcal S_{s_j})\right).
  \label{eq:retrieval-score}
\end{equation}
We fill a shared budget by alternating between the entities' rankings; no
retriever is trained and no query is decomposed. We evaluate on
Dream-7B-Instruct and LLaDA-8B-Instruct.

The benchmark is built from PopQA facts~\citep{mallen2023popqa} and is
designed around relations: $5{,}528$ evidence cards over five domains and
$19$ relations, every entity carrying exactly the four relations of its
domain, searched globally for $10{,}800$ test queries balanced over the
domains and over three layouts, which name one entity, two entities with
adjacent entity--relation clauses, or two entities with crossed references
(``$A$ and $B$: the director of the former and the composer of the
latter''). Single R@1 scores the correct card; dual-target All@2 requires
both correct cards within two retrieved cards. A second, entity-conditioned
protocol hands every method the same entity-linked candidates, four cards
per entity, so that only the relation has to be chosen. Baselines are BM25,
whole-query pooling on the same backbones, DiffRetriever's parallel
representative tokens~\citep{wang2026diffretriever}, and PromptReps on
Qwen2.5-7B-Instruct~\citep{zhuang2024promptreps}. Construction, readout
selection and protocols are in App.~\ref{app:retrieval}.

\subsection{Results}
\label{sec:retrieval:results}

Table~\ref{tab:retrieval-popqa} reports both protocols. On single-entity
queries every method retrieves well. On two-entity queries every
single-vector method collapses: BM25, pooling, DiffRetriever and PromptReps
retrieve both cards for at most $3.06\%$ of adjacent and $2.50\%$ of crossed
queries, because one vector cannot hold two entity--relation bindings.
Per-entity access states retrieve both for $38.28$--$41.94\%$ of adjacent and
$21.78$--$38.08\%$ of crossed queries from the selected layers. The
entity-conditioned protocol, which removes entity coverage as a factor,
leaves the ordering unchanged ($41.94/38.08$ against at most $16.35/16.26$):
the advantage lies in choosing the relation within an entity's cards, which
is what the state binds.

\begin{table}[htbp]
\centering
\footnotesize
\setlength{\tabcolsep}{3pt}
\caption{\textbf{Controlled benchmark: global search and entity-conditioned
retrieval} (\%). All methods search the same $5{,}528$ cards for the same
$10{,}800$ test queries. Single is R@1; Adj./Cross.\ are dual-target All@2.
Time is measured in ms/query. Shaded cells mark the best value in each column.}
\label{tab:retrieval-popqa}

\resizebox{\linewidth}{!}{%
\begin{tabular}{lrrrrrrrr}
\toprule
& \multicolumn{3}{c}{\textbf{Global corpus}}
& \multicolumn{3}{c}{\textbf{Entity-conditioned}}
& \multicolumn{2}{c}{\textbf{Time}~\metdn} \\
\cmidrule(lr){2-4}\cmidrule(lr){5-7}\cmidrule(lr){8-9}
Method
& Single & Adj. & Cross.
& Single & Adj. & Cross.
& Global & Entity \\
& R@1~\metup & All@2~\metup & All@2~\metup
& R@1~\metup & All@2~\metup & All@2~\metup
& \scriptsize ms/query & \scriptsize ms/query \\
\midrule
\multicolumn{9}{@{}l}{\textit{Shared baselines}} \\
BM25
& 92.82 & 1.38 & 1.01 & 95.35 & 11.83 & 11.83
& \best{0.51} & \best{0.31} \\
Qwen2.5 pooling
& 69.06 & 1.28 & 0.00 & 81.19 & 8.83 & 8.14
& 12.76 & 0.43 \\
PromptReps (20/20), hybrid
& 92.53 & 1.58 & 1.56 & 93.47 & 16.35 & 16.26
& 264.98 & 263.20 \\
\midrule
\multicolumn{9}{@{}l}{\textit{Dream-7B-Instruct}} \\
Whole-query pooling
& 83.53 & 3.06 & 0.58 & 87.03 & 13.67 & 11.92
& 13.41 & 13.26 \\
DiffRetriever (4/16), dense
& 80.03 & 0.89 & 0.25 & 86.17 & 9.81 & 9.42
& 14.27 & 13.87 \\
DiffRetriever (4/16), hybrid
& 97.14 & 1.25 & 1.28 & \best{99.17} & 7.28 & 6.69
& 17.52 & 16.73 \\
Subject (L0)
& 78.50 & 24.44 & 13.33 & 78.61 & 24.56 & 13.42
& 1.88 & 1.47 \\
Subject (L27)
& 81.01 & 38.28 & 21.78 & 81.03 & 38.28 & 21.76
& 13.64 & 13.29 \\
\midrule
\multicolumn{9}{@{}l}{\textit{LLaDA-8B-Instruct}} \\
Whole-query pooling
& 88.25 & 2.11 & 2.50 & 90.08 & 9.72 & 10.56
& 20.68 & 20.37 \\
DiffRetriever (4/16), dense
& 76.36 & 0.61 & 0.67 & 82.53 & 10.64 & 10.11
& 21.72 & 21.05 \\
DiffRetriever (4/16), hybrid
& \best{97.69} & 1.36 & 1.03 & 98.61 & 9.19 & 8.78
& 25.88 & 24.88 \\
Subject (L0)
& 68.03 & 13.32 & 9.44 & 68.42 & 13.61 & 9.78
& 2.00 & 1.45 \\
Subject (L31)
& 89.08 & \best{41.94} & \best{38.08}
& 89.08 & \best{41.94} & \best{38.08}
& 20.97 & 20.35 \\
\bottomrule
\end{tabular}%
}
\end{table}
\paragraph{The state is usable as soon as it is formed.}
Read after the first block, the access state already retrieves both cards
for $24.44\%/13.33\%$ of adjacent/crossed queries on Dream, and the deeper
layers refine it to $38.28/21.78$ (LLaDA: $13.32/9.44 \to 41.94/38.08$). The
readout carries to natural text (App.~\ref{app:retrieval-natural}): on the
comparison questions of HotpotQA and 2WikiMultiHopQA, the first-block
readout returns evidence from both gold paragraphs (Both@2) for $41.82\%$
and $37.37\%$ of questions against $25.15\%/26.93\%$ for BM25, at $2.6$\,ms per query against
$29$--$93$\,ms for full-depth encoders, and a frozen reader given that
evidence gains $4$--$8$ F1 points over the strongest baseline. The early
readout is also cheap: reading the state at L0 costs $1.5$--$2.0$\,ms per
query on both backbones, against $13$--$26$\,ms for the full-depth neural
retrievers and $265$\,ms for PromptReps (Table~\ref{tab:retrieval-popqa}).

\paragraph{The retrieval behaviour is causal.}
Subtracting the first attention layer's relation-token messages at the
subject positions drops within-entity relation accuracy to the four-way
chance level in all three layouts while entity identification stays at
$100\%$, and substituting the messages of a different relation redirects
retrieval toward that relation (App.~\ref{app:retrieval-causal}). Reading
the state is thus the counterpart of the write in \S\ref{sec:method}: the
layer-$0$ routing that an edit must cover is what a retrieval vector taken
from the state already contains.


\section{Conclusion}
\label{sec:conclusion}

We asked what a masked DLM computes when it accesses a fact,
before asking how to change or use it.
Causal tracing reveals a two-stage mechanism: layer-0 attention
routes relation wording into the visible entity, and early MLPs
look up the fact on the resulting access state.
This state depends on the relation's surface form and query
direction, rather than on the entity alone.
Writing to this state, \methodname{} fits shared values across
multiple query views, improving paraphrase and reverse
generalisation while preserving source-prompt reliability
and locality.
Reading it provides a complementary insight: per-entity
access states can serve as relation-specific retrieval vectors.
Our experiments offer a proof of concept for using these
states to access external evidence, connecting knowledge
editing and retrieval through a shared mechanism of
factual access.
Two limitations bound these claims. The editor assumes identifiable
subject, relation and object spans, which CounterFact and BAKE supply but
arbitrary user statements may not, and free-generation transfer on QAEdit
remains incomplete, since wrong answers are still the majority on
paraphrased questions (App.~\ref{app:limitations}).

\section*{AI Use Statement}

Generative AI models were used as experimental components, as described in the paper. AI-assisted tools were otherwise used only for limited language polishing and code troubleshooting. All methodological decisions, experimental results, interpretations, and final manuscript content were reviewed and approved by the authors, who take full responsibility for this work.

\section*{Reproducibility Statement}

The paper specifies the mask-native causal tracing protocol, multi-view editing procedure, and per-entity retrieval readouts. Appendix~\ref{app:causal-extra} reports additional mechanism and role-reversal checks. Appendix~\ref{app:hyperparams} records editing hyperparameters, context templates, seed, precision, and compute, while Appendix~\ref{app:update-details} details covariance estimation under the DLM masking distribution. Appendix~\ref{app:editing-details} documents editing benchmarks, baselines, and ablations. Appendix~\ref{app:retrieval} specifies corpus construction, validation and test splits, readout selection, scoring, and evaluation protocols for both controlled and natural retrieval settings. Seed $42$ fixes sampled masks and covariance estimation; the editor uses no random initialisation or stochastic retraining. Code is available at \url{https://github.com/Guoyuchen1/dlm-factual-access}.

\section*{Ethics Statement}

Knowledge editing can be misused to make false claims accessible through multiple phrasings and query directions. Our experiments use controlled factual-edit benchmarks and report locality and neighborhood specificity to assess unintended changes; these results do not establish safety for deployment. The retrieval experiments use public benchmark corpora and may inherit errors or biases in their source material. We collected no private user data and conducted no human-subject experiments.


\IfFileExists{iclr2025_conference.bib}{%
  \bibliography{iclr2025_conference}
  \bibliographystyle{iclr2025_conference}
}{}

\appendix

\section{Detailed Related Work}
\label{app:related-work}

\subsection{Knowledge editing in autoregressive language models}

Knowledge editing aims to change selected model behaviour while preserving
unrelated predictions. Existing approaches include constrained fine-tuning and
learned editors \citep{zhu2020modifying,decao2021editing,mitchell2022mend},
direct parameter rewriting \citep{meng2022rome,meng2023memit,li2024pmet,fang2025alphaedit},
and external or semi-external memories designed for scalable and lifelong
updates \citep{mitchell2022serac,zheng2023ike,hartvigsen2023grace,wang2024wise}.
The locate-then-edit family is closest to our work. ROME models an MLP as a
key--value memory and installs one association with a rank-one update;
MEMIT distributes related updates across several layers. Later methods refine
the edited subspace or update rule, but retain the AR factorisation in which a
subject-side representation is read from a prefix and the target is predicted
to its right.

This literature has traditionally evaluated three properties: reliability on
the source prompt, generalisation to paraphrases, and locality on unrelated
prompts. More recent benchmarks expose dimensions not captured by this triad.
MQuAKE and RippleEdits test whether a local change propagates to related or
multi-hop consequences \citep{zhong2023mquake,cohen2024rippleedits}; QAEdit
shows that edited-prompt success can degrade under natural questions, free-form
decoding, and sequential editing \citep{yang2025mirage}. Our experiments
therefore include both conventional forced-choice scores and open-ended
statement-to-question transfer.

\subsection{From prompt editing to statement-level editing}

Most locate-then-edit methods receive a natural-language statement but reduce it
to one directional conditional. For a fact $(s,r,o)$, an AR editor typically
optimises the association queried by $p(o\mid s,r)$. This is sufficient for the
source continuation, but it is not equivalent to making the model use the
statement as knowledge. The same proposition may be queried by masking the
object, recovering the subject from $(r,o)$, judging whether the proposition is
true, omitting part of the relation, or embedding it in a question and
generating an answer. Under left-to-right attention, several of these
conditionals cannot be formed from the same token order because the desired
target appears before some of its evidence.

Our use of \emph{statement-level editing} refers to this distinction. The user
still supplies one declarative edit, not a hand-written set of questions or
paraphrases. The editor then uses the DLM's native corruption operator to derive
multiple conditional views of that statement. This differs from ordinary prompt
augmentation in two ways. First, the views include different prediction
directions, not only surface rewrites of the same forward prompt. Second, they
are generated from the model's masking distribution rather than from the
evaluation benchmark's paraphrase set. Consequently, success on a held-out
question or reverse probe tests whether the installed association is reachable
from a new conditional view, rather than whether the exact evaluation prompt
was used during editing.

This framing should not be read as claiming that one forward update
automatically induces all reverse knowledge. Our BAKE ablation shows the
opposite: bidirectional attention makes an object-side conditional
\emph{expressible}, but most reverse gains arise only after the reverse
direction is explicitly included in the edit group. The contribution is that
both directions, relation occlusions, and contextual variants can be derived
from one statement and installed in a common early-MLP block.

\subsection{Reversal, bidirectional knowledge, and model editing}

The reversal curse describes the failure of AR models trained on ``$A$ is
$B$'' to reliably answer the inverse query ``$B$ is $A$''
\citep{berglund2023reversal}. The factorisation account attributes this failure
to which tokens are predicted from which contexts during language modelling
\citep{kitouni2024factorization}. BAKE transfers the issue from training to
model editing: an editor may successfully change the forward association while
leaving reverse retrieval unchanged. BIRD introduces bidirectional objectives
within the AR setting \citep{ma2023bird}, but AR token order still constrains
which conditionals can be represented as ordinary target-final prompts.

Masked DLMs reconstruct arbitrary masked subsets and therefore weaken this
structural asymmetry \citep{sahoo2024mdlm,nie2025llada,ye2025dream}. Recent
analyses nevertheless find that bidirectional attention does not eliminate
reversal failures; proposed explanations include entity fragmentation,
asymmetry in data or relations, and the degree to which forward and reverse
queries share representations \citep{he2026differ,jeon2026theoretical}. These
findings motivate our separation between \emph{availability} and
\emph{reachability}: a reverse conditional can exist in the architecture while
a parameter update fitted to one key still fails to activate from the reverse
query. \methodname{} addresses reachability by constructing and installing
keys from both sides of the same statement.

\subsection{Knowledge editing in masked diffusion language models}

Concurrent work has established that factual associations can be edited in
masked diffusion language models. \citet{park2026editing} transfer
locate-then-edit to masked DLMs and show that the edit \emph{site} carries over
from the AR setting: causal tracing highlights an early-to-mid-layer MLP at the
last subject token, and editing is most effective there. They then identify a
failure mode specific to diffusion decoding: a value fitted on a fully masked
target degrades once some target tokens have been revealed, because the edit
was never optimised for those intermediate states. They correct it by exposing
partially unmasked target states during value optimisation. TimeROME-DLM
instead combines temporal causal tracing with a low-rank residual memory
applied at inference time, leaving the backbone parameters unchanged
\citep{yao2026timerome}.

\paragraph{Two orthogonal failure modes.}
We take the localisation of \citet{park2026editing} as given and ask a
different question about the same site: what the retrieval step there actually
reads. The two diagnoses concern different axes of the same editor. Theirs is
about the \emph{decoding trajectory}: a value optimised on one masking state of
the target is not optimised for the intermediate states traversed while that
target is generated. Ours is about the \emph{query entrance}: a value optimised
at one key is only weakly activated when the relation is reworded or the query
is posed from the object side (\eqref{eq:binding}). Neither subsumes the other,
and they compose: in the notation of \S\ref{sec:method:views} their correction
is a further view type supervised on the same shared value, a partial reveal of
$\mathcal{S}_o$ rather than an occlusion of $\mathcal{S}_r$. App.~\ref{app:compose-parkjo} carries this composition out on QAEdit and
confirms it: adding their partial-reveal views to our view family raises the
weighted score on LLaDA-8B from $0.250$ to $0.430$ on source prompts and from
$0.187$ to $0.352$ on paraphrased questions, and largely removes the
multi-token degradation of App.~\ref{app:qaedit-length}. Their method is
therefore complementary to ours rather than a competing one.

\paragraph{Why no concurrent DLM editor appears in Table~\ref{tab:bake}.}
The reason is structural rather than empirical. Both concurrent methods
supervise a forward conditional in which the subject stays visible and the key
is read on the subject side; neither installs an object-side association, so
neither can make a forward edit reachable from a reverse query, which is the
quantity the reverse columns of Table~\ref{tab:bake} measure. The
\textsc{Single-Forward} rows already bound what any forward-only construction
achieves there. TimeROME-DLM is further removed, since it leaves the backbone
unchanged and edits through an inference-time residual memory, which is not
commensurable with a key-coverage account of a weight update. We therefore use constrained
fine-tuning (FT+L) as a reproducible masked-DLM baseline and, more importantly,
compare \methodname{} against matched single-view and single-direction variants
that isolate the contribution of each proposed design axis. Published AR
results are collected in App.~\ref{app:ar-refs} as scale references rather
than controlled comparisons.
\begin{table}[htbp]
\centering
\footnotesize
\setlength{\tabcolsep}{3.5pt}
\renewcommand{\arraystretch}{1.1}
\caption{\textbf{Positioning of closely related editing settings.}
Existing masked-DLM editors primarily address robustness across diffusion
states or introduce inference-time memories. \methodname{} instead expands one
declarative statement into a family of conditional retrieval views.}
\label{tab:related-positioning}
\begin{tabular}{L{0.155\linewidth}L{0.095\linewidth}L{0.165\linewidth}L{0.28\linewidth}L{0.115\linewidth}}
\toprule
\textbf{Setting} & \textbf{Backbone} & \textbf{Update mechanism}
  & \textbf{Variation explicitly covered} & \textbf{Reverse edit} \\
\midrule
ROME / MEMIT & AR & Backbone MLP weights
  & Context templates around one forward conditional & Not native \\
\addlinespace[2pt]
BIRD / BAKE & AR & Bidirectional objectives / evaluation
  & Forward--reverse objectives under AR factorisation & Explicit \\
\addlinespace[2pt]
Park \& Jo & Masked DLM & Early--middle MLP weights
  & Partially unmasked target states during decoding & Not central \\
\addlinespace[2pt]
TimeROME-DLM & Masked DLM & Inference-time residual memory
  & Diffusion-time / temporal retrieval states & Not central \\
\addlinespace[2pt]
\methodname{} & Masked DLM & Early-MLP down-projections
  & Relation form, bidirectional context, and forward--reverse views from one statement
  & Explicit \\
\bottomrule
\end{tabular}
\end{table}

\subsection{Reading language-model representations for retrieval}
\label{app:related-retrieval}

Dense retrieval encodes a query and a passage into one vector each and ranks
by inner product~\citep{karpukhin2020dpr}; late-interaction models keep one
vector per token and score with MaxSim~\citep{khattab2020colbert}; BM25 ranks
by lexical overlap~\citep{robertson2009bm25}. A second line uses a frozen
language model as the encoder. PromptReps prompts an autoregressive LM to emit
representative tokens whose hidden states serve as dense and sparse
representations~\citep{zhuang2024promptreps}; LLM2Vec and Echo embeddings
obtain a pooled representation from a decoder by enabling bidirectional
attention or repeating the input~\citep{behnamghader2024llm2vec,springer2024echo};
DiffRetriever carries the representative-token recipe to masked diffusion LMs,
appending mask positions to query and passage and encoding them in
parallel~\citep{wang2026diffretriever}. In all of these the query becomes one
vector, or a small set of vectors placed by a prompt, rather than a
representation read from the model's own processing of the entities the
query mentions. Entity-centric questions are a known weak point of pooled
dense retrievers~\citep{sciavolino2021entity}, and PopQA, from which our
evidence corpus is built, isolates the long-tail entity facts on which
retrieval augmentation is most needed~\citep{mallen2023popqa}. Questions that
ask about several entities, such as the comparison questions of HotpotQA and
2WikiMultiHopQA~\citep{yang2018hotpotqa,ho2020twowiki}, are usually handled
upstream, by decomposing the question into sub-questions before
retrieval~\citep{press2022selfask,khot2023decomposed} or by entity linking
followed by per-entity retrieval.

Our retriever differs on both counts. It reads representations the model
already computes, the access state at each entity position, so the
per-entity, relation-bound query vectors that decomposition methods construct
externally are obtained from the contextualised entity positions of the query
itself, with no retriever training, no prompt-placed representative tokens and
no parser. Because the state is present from the first block on
(\S\ref{sec:mechanism}), the query encoder can read it early, and because the
same state is what the early MLPs read for parametric lookup, the retrieval
readout and the editing write of \S\ref{sec:method} operate on one
representation. Benchmarks, protocols and controls are reported in App.~\ref{app:retrieval}.

\section{Additional Mechanism Analyses}
\label{app:causal-extra}
\label{app:mechanism-robustness}

This appendix reports five auxiliary causal-tracing analyses that complement
\S\ref{sec:mechanism}. The first three characterise the retrieval pipeline on
LLaDA-8B in more detail and justify claims made in passing in the main text:
(i)~the early peak is dominated by MLP rather than attention
(App.~\ref{app:selective-ablation}), (ii)~the two visual ``peaks'' are
arranged in a strict retrieve-then-decode \emph{series}, not in parallel
(App.~\ref{app:cross-peak}), and (iii)~the causal locus migrates from subject
tokens to the \texttt{[MASK]} position only in the last few layers
(App.~\ref{app:mask-mirror}). The last two establish that the pipeline is not
an artefact of one checkpoint or one mask layout: it replicates on Dream-7B
(App.~\ref{app:dream-repro}) and runs role-swapped when the subject rather
than the object is masked (App.~\ref{app:mask-subject}), which is the
property that Definition~\ref{def:access-state} and the reverse views of
\S\ref{sec:method} rely on. Unless otherwise stated, all numbers are on
CounterFact with LLaDA-8B and follow the tracing protocol of
\S\ref{sec:tracing}.

\begin{figure}[htbp]
  \centering
  \safeincludegraphics[width=0.98\linewidth]{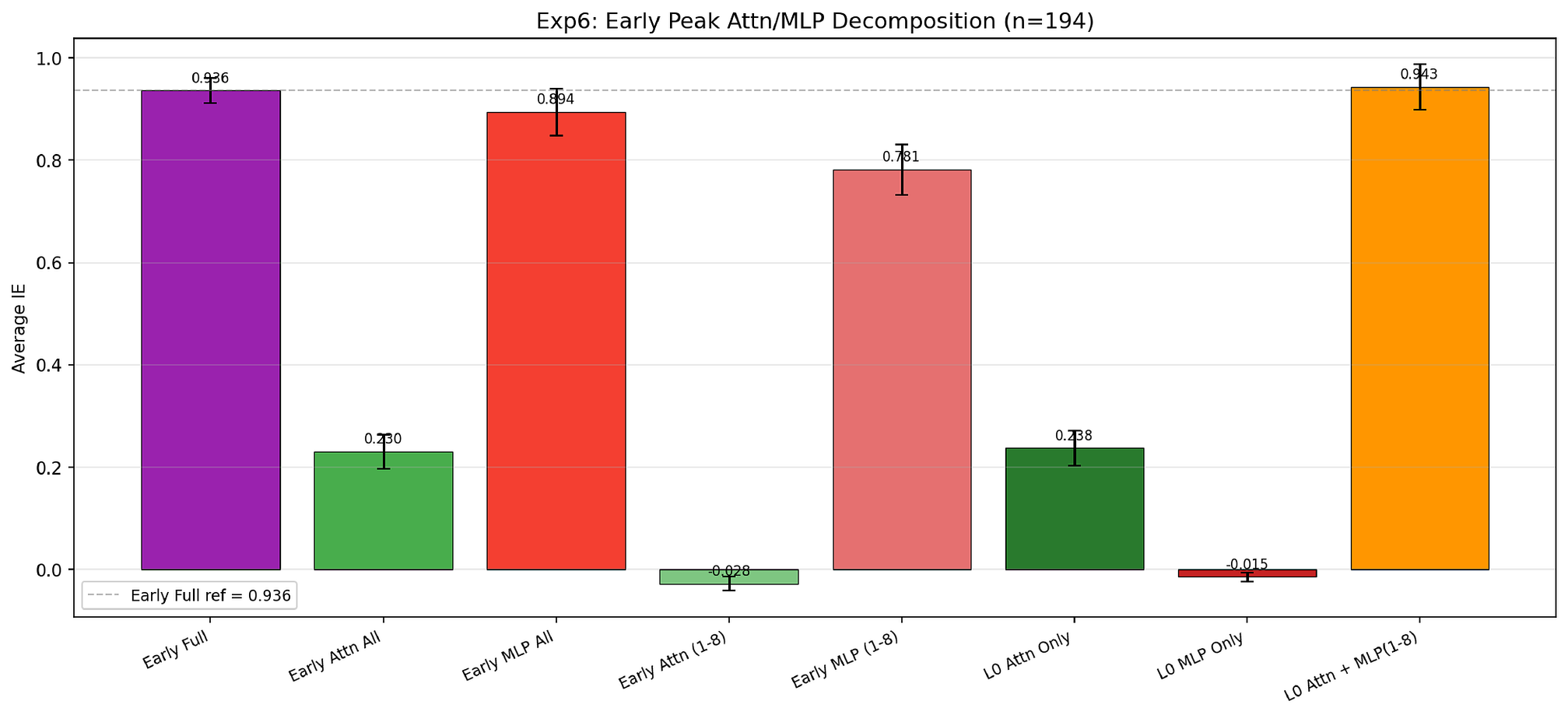}
  \caption{\textbf{Early-block decomposition} (referenced in
  \S\ref{sec:footprint}). IE recovered by restoring different subsets of the
  early block. Layer-$0$ attention combined with MLPs $1$--$8$ ($0.943$) matches
  the full early block ($0.936$), while either sub-component alone falls
  short.}
  \label{fig:early-decomp}
\end{figure}

\subsection{Selective Peak Ablation \& Channel Decomposition}
\label{app:selective-ablation}

\begin{figure}[htbp]
  \centering
  \safeincludegraphics[width=0.98\linewidth]{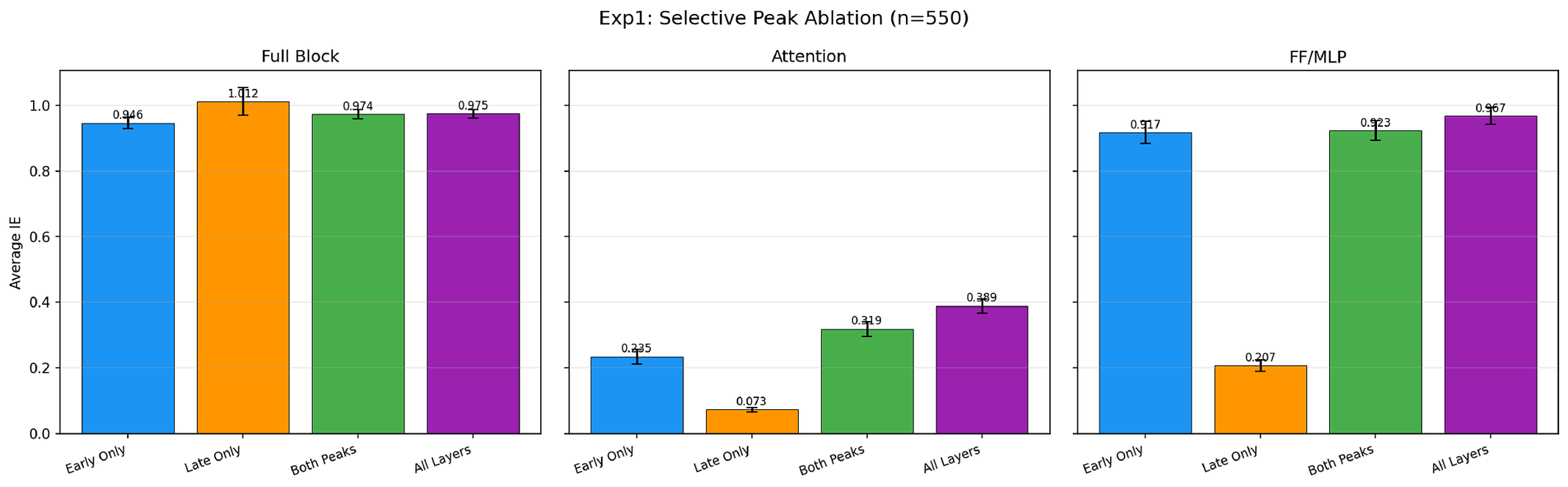}
  \caption{\textbf{Selective peak ablation with channel decomposition
  ($n{=}550$).} Restoring the early peak alone already recovers most of
  the causal effect on the MLP channel ($0.917$ vs.\ $0.967$ for all
  layers), while restoring the late peak alone leaves the attention
  channel almost silent ($0.073$).}
  \label{fig:app-exp1}
\end{figure}

We compare four restoration policies applied to subject tokens under the
mask-corruption protocol of \S\ref{sec:tracing}:
\emph{Early Only} (restore layers $0$--$8$),
\emph{Late Only} (restore layers $18$--$25$),
\emph{Both Peaks} (restore both windows), and
\emph{All Layers} (restore the entire stack). For each policy we further
split the restored signal by channel into \emph{Full Block},
\emph{Attention Only}, and \emph{FF/MLP Only}
(Fig.~\ref{fig:app-exp1}, $n{=}550$).

\paragraph{Findings.}
Three observations follow directly from Fig.~\ref{fig:app-exp1}.
\emph{First}, on the Full Block the four policies are nearly
indistinguishable ($0.946$--$1.012$), which by itself would suggest that
either peak is sufficient. \emph{Second}, this appearance is entirely
carried by MLPs: the MLP-only panel shows \emph{Early Only} $=0.917$
compared with \emph{Late Only} $=0.207$, a $\sim 4.4\times$ gap.
\emph{Third}, the attention channel is uniformly weak (all four
conditions $\le 0.389$), consistent with the single-layer analysis in
\S\ref{sec:stacking} that attributes essentially all attention-side causal
mass to layer $0$.

\subsection{Cross-Peak Patching: Are the Two Peaks in Series or in Parallel?}
\label{app:cross-peak}

Sec.~\ref{app:selective-ablation} shows that either peak, restored alone
on the Full Block, produces near-ceiling IE. This is compatible with two
very different mechanisms: the peaks could be
\emph{redundant parallel channels} (each independently reconstructs the
answer), or a \emph{serial pipeline} (the late peak only works because the
early peak feeds it). We disambiguate by conditional patching: we restore
one peak while simultaneously \emph{blocking} the other, i.e.\ pinning the
blocked peak's hidden states to their corrupted values so that no
information can flow through it.

\begin{figure}[htbp]
  \centering
  \safeincludegraphics[width=0.98\linewidth]{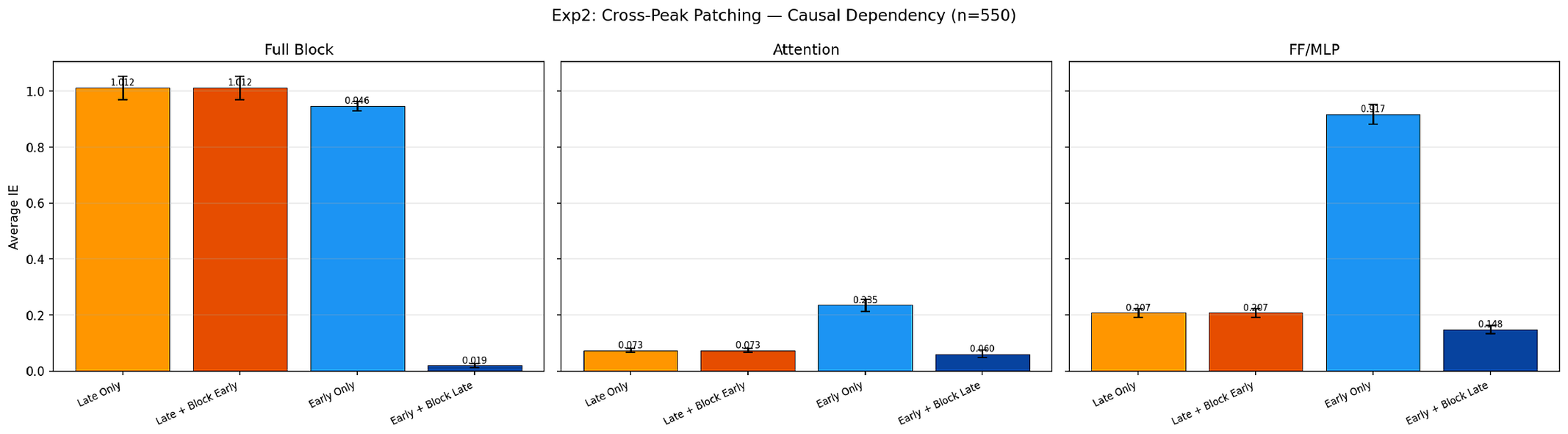}
  \caption{\textbf{Cross-peak conditional patching ($n{=}550$).}
  \emph{Late $+$ Block Early} preserves the late peak's IE
  ($1.012 \to 1.012$ on Full Block), whereas \emph{Early $+$ Block Late}
  destroys the early peak's effect ($0.946 \to 0.019$). The asymmetry
  is strongest on the MLP channel ($0.917 \to 0.148$).}
  \label{fig:app-exp2}
\end{figure}

\paragraph{Findings.}
The asymmetry in Fig.~\ref{fig:app-exp2} is striking:
\begin{itemize}[leftmargin=1.4em,itemsep=0.15em,topsep=0.15em]
  \item \emph{Late Only} $=1.012$ and \emph{Late $+$ Block Early} $=1.012$
        (Full Block): blocking the early peak leaves the late peak's
        performance untouched, because when we \emph{restore} the late
        peak we hand it a clean representation directly.
  \item \emph{Early Only} $=0.946$ but \emph{Early $+$ Block Late}
        $=0.019$: restoring the early peak alone is useless if the late
        peak is blocked, because the recovered representation cannot
        reach the readout.
\end{itemize}
The MLP channel shows the same pattern with an even larger gap
($0.917 \to 0.148$).

\paragraph{Take-away.}
The two peaks are \emph{serially composed}, not parallel. The early
peak is the \emph{source} of the fact-relevant signal; the late peak is a
downstream \emph{conduit} that carries the already-retrieved
representation toward the \texttt{[MASK]} readout. This is exactly what
one expects from the \emph{retrieve-then-decode} view articulated in
\S\ref{sec:bimodal}.

\subsection{Subject-to-\texttt{[MASK]} Handoff: Mirror Analysis on the Readout Position}
\label{app:mask-mirror}

\begin{figure}[htbp]
  \centering
  \safeincludegraphics[width=0.98\linewidth]{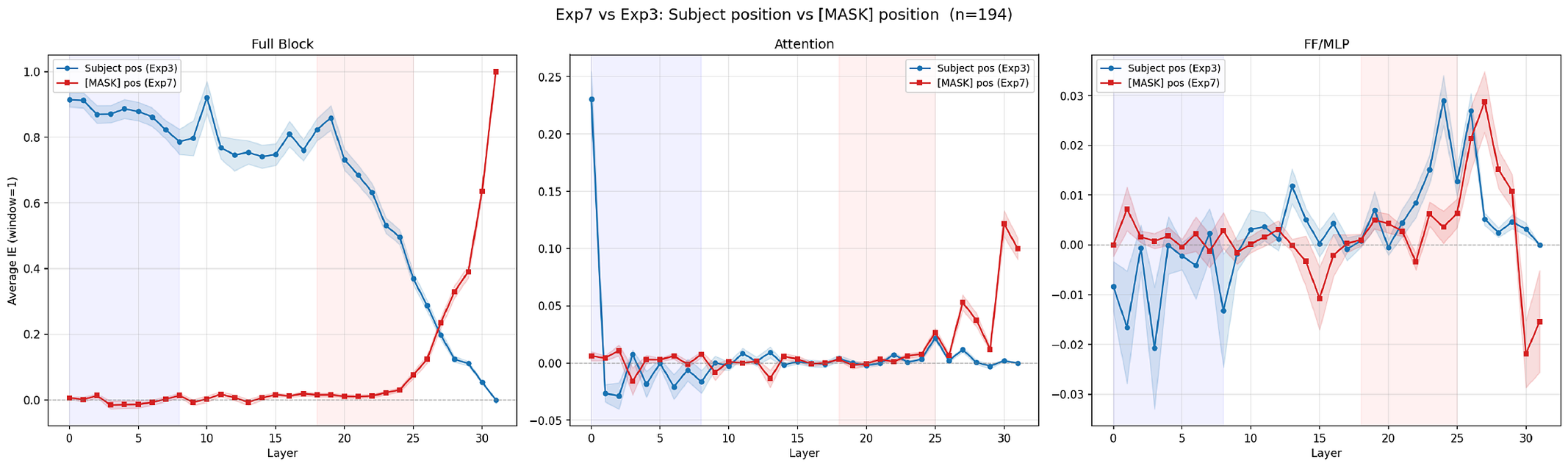}
  \caption{\textbf{Subject position (Exp~3) vs.\ \texttt{[MASK]} position
  (Exp~7), $n{=}194$.} Left: full-block IE per layer.
  Middle / Right: attention-only and MLP-only splits. The two positions
  are almost mutually exclusive as causal loci: subject-side dominance
  gives way to \texttt{[MASK]}-side dominance only in the last few
  layers.}
  \label{fig:app-exp7}
\end{figure}

The analyses so far restore hidden states at \emph{subject} tokens. To see
how and when the signal migrates to the position that actually emits the
prediction, we run a mirror experiment (Exp~7) that keeps the tracing
protocol identical to Exp~3 (\S\ref{sec:stacking}) but restores hidden
states at the \texttt{[MASK]} position instead of at the subject range.
Fig.~\ref{fig:app-exp7} overlays the two curves.

\paragraph{Findings.}
Reading Fig.~\ref{fig:app-exp7} from left to right along the layer axis
gives a four-stage picture of fact recall in a masked DLM:
\begin{enumerate}[leftmargin=1.6em,itemsep=0.15em,topsep=0.15em]
  \item \textbf{Layer $\sim\!0$: subject-side anchoring by attention.}
        The attention panel shows a single dominant spike at layer $0$
        on the \emph{subject} curve ($\approx\!0.23$), with the
        \texttt{[MASK]} curve near zero. This is the context-fusion
        step characterised in \S\ref{sec:l0-surgery}.
  \item \textbf{Layers $\sim\!1$--$24$: subject-side persistence and
        processing.} On the Full Block the subject curve stays in
        $[0.7, 0.9]$ while the \texttt{[MASK]} curve remains near zero,
        indicating that the fact-relevant representation continues to
        live \emph{at the subject positions} throughout most of the
        network.
  \item \textbf{Layers $\sim\!25$--$28$: subject $\to$ \texttt{[MASK]}
        handoff.} The two curves cross: the subject curve drops sharply
        while the \texttt{[MASK]} curve rises. Bidirectional attention
        allows this transfer to happen in a narrow band rather than
        being smeared across many layers.
  \item \textbf{Top layers: \texttt{[MASK]}-side local readout.}
        In the last few layers the \texttt{[MASK]} curve dominates
        (Full Block $\to 1.0$; attention $\approx\!0.12$; MLP peaks
        around layer $27$). The final decoding is performed locally at
        the readout position, after the retrieved representation has
        been delivered there.
\end{enumerate}

\subsection{Cross-Model Robustness: Dream-7B Reproduces the LLaDA Pipeline}
\label{app:dream-repro}

We rerun the 1-D causal trace of \S\ref{sec:tracing} on
\textbf{Dream-7B}, an independently trained masked DLM whose
architecture, training data and mask schedule differ from LLaDA-8B.
For each fact we compute the average indirect effect
$\mathrm{AIE}(\ell, \tau)$ separately at each layer $\ell$ and each
role-typed position $\tau \in \{\text{Subject}, \text{Relation},
\text{Prefix}, \text{[MASK]}\}$.

\begin{figure}[htbp]
  \centering
  \safeincludegraphics[width=0.98\linewidth]{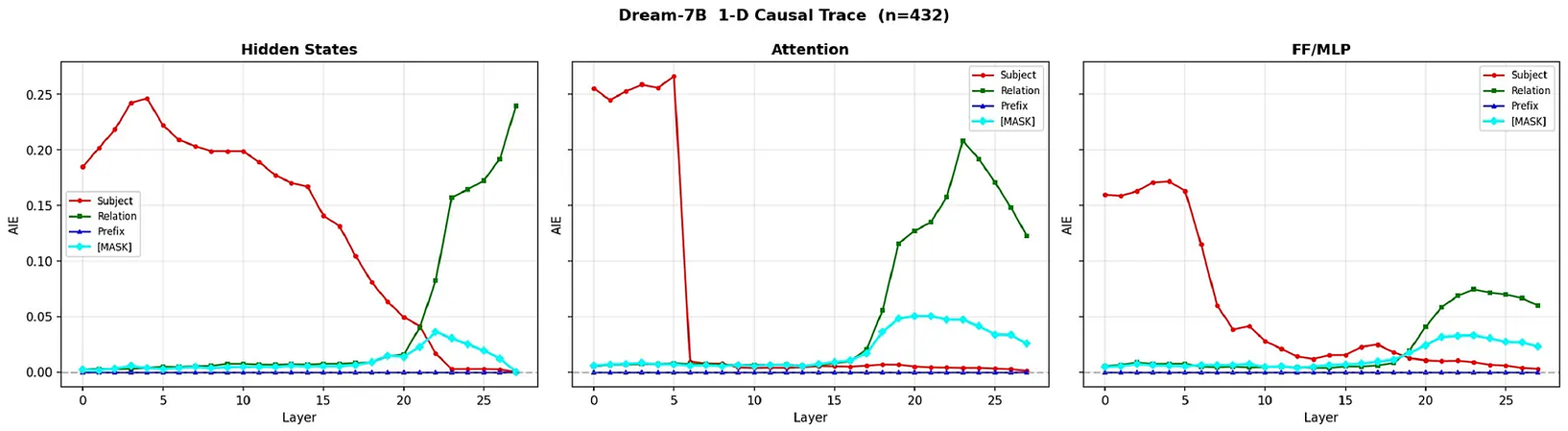}
  \caption{\textbf{Dream-7B 1-D causal trace, $n{=}432$.}
  Left: hidden states. Middle: attention only. Right: FF/MLP only.
  Four curves correspond to the four role-typed positions. Dream-7B
  reproduces every qualitative feature reported on LLaDA-8B: a
  subject-anchored early peak in Hidden and MLP, a very large
  low-layer spike on the \emph{Attention} channel at the subject
  positions (layers $0$--$5$, $\mathrm{AIE}\!\approx\!0.25$), a
  Relation-side attention peak around layers $18$--$23$, and a
  \texttt{[MASK]}-side takeover in the top layers.}
  \label{fig:dream-1d}
\end{figure}

\paragraph{Findings.}
Fig.~\ref{fig:dream-1d} matches the LLaDA fingerprint on every axis we
tested:
\begin{enumerate}[leftmargin=1.6em,itemsep=0.15em,topsep=0.15em]
  \item \emph{Low-layer subject-side attention dominates.} On the
        attention panel, the Subject curve reaches
        $\mathrm{AIE}\!\approx\!0.25$ at layers $0$--$5$ and then
        collapses to near zero from layer $6$ onwards, mirroring the
        single-layer-$0$ dominance we established on LLaDA
        (\S\ref{sec:stacking}).
  \item \emph{Subject-anchored bimodal footprint on Hidden / MLP.} The
        Subject curve on Hidden states peaks around layer $5$ and
        remains elevated up to layer $\sim\!20$; the MLP panel shows
        the same subject-side elevation over the early block.
  \item \emph{Relation-side second peak in the top block.} An
        attention-side Relation peak appears at layers $18$--$23$
        ($\mathrm{AIE}\!\approx\!0.20$), consistent with the
        subject$\to$readout handoff band identified on LLaDA
        (App.~\ref{app:mask-mirror}).
  \item \emph{\texttt{[MASK]}-side local readout.} The \texttt{[MASK]}
        curve stays near zero for most of the network and rises only
        in the last few layers, exactly as observed for LLaDA.
\end{enumerate}

Because Dream-7B shares neither pretraining data nor exact model shape
with LLaDA-8B yet reproduces all four features, \textbf{the ``low-layer
context fusion $+$ early-MLP lookup'' pipeline appears to be shared across
the two masked DLMs studied rather than specific to the LLaDA checkpoint}. This is the mechanistic version of the empirical
cross-model transfer already reported in \S\ref{sec:experiments}, and
it is what makes it safe to apply the same Multi-View recipe unchanged
across the two DLMs. It does not by itself explain why the editor's
\emph{quantitative} gains differ so much between the two checkpoints
(App.~\ref{app:limitations}).

\subsection{Cross-Layout Robustness: The Mechanism Survives a Subject-Side Mask}
\label{app:mask-subject}

The tracing setup in \S\ref{sec:tracing} always masks the object span at
the \emph{tail} of the prompt. This raises a legitimate concern that the
pipeline we characterise is specific to tail-side prediction, and that
observed ``fact retrieval'' merely reflects how DLMs handle their
canonical next-mask completion. To rule this out we construct a
\emph{subject-side} variant in which we \textbf{keep the object visible}
and instead \textbf{mask the subject span in the middle of the prompt}
(denoted \texttt{[MASK\_subj]}). We then re-run the 1-D causal trace at
each role-typed position, using LLaDA-8B as the reference model.

\begin{figure}[htbp]
  \centering
  \safeincludegraphics[width=0.98\linewidth]{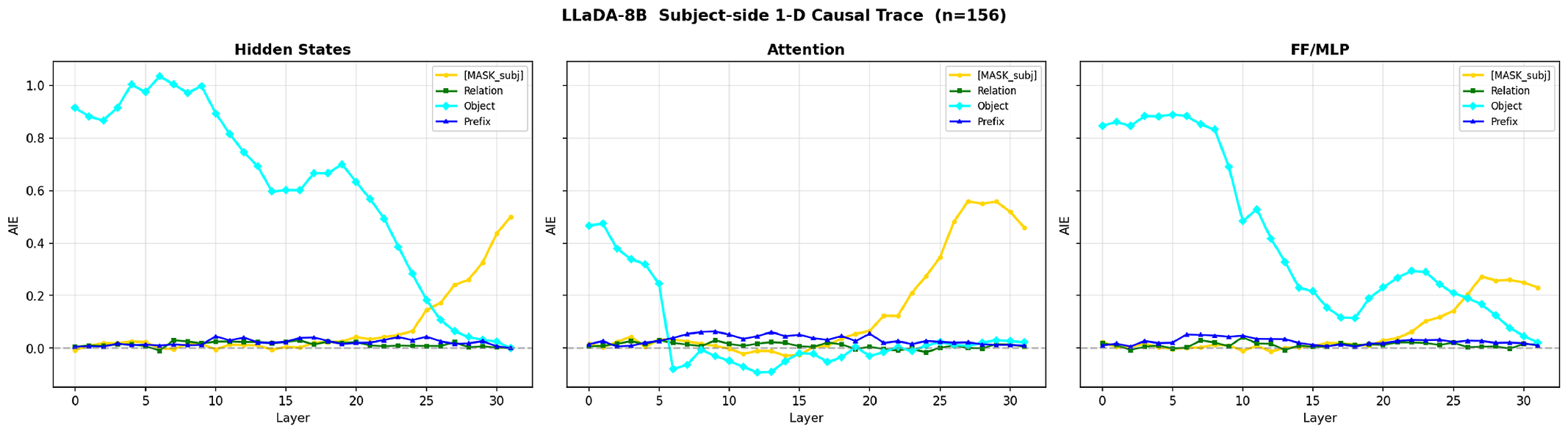}
  \caption{\textbf{LLaDA-8B subject-side 1-D causal trace, $n{=}156$.}
  Setting: the object is kept clean; the subject span is replaced by
  \texttt{[MASK\_subj]}. Curves show AIE at each role-typed position
  across layers. The \textbf{Object} position now plays the causal role
  that the subject played in the tail-mask setting: it dominates
  Hidden / MLP in the early-to-mid layers and its attention spikes at
  the lowest layers. The \texttt{[MASK\_subj]} position takes over in
  the top layers, exactly mirroring the \texttt{[MASK]} takeover of
  App.~\ref{app:mask-mirror}.}
  \label{fig:mask-subject}
\end{figure}

\paragraph{Findings.}
Compared with the tail-mask setting, Fig.~\ref{fig:mask-subject} shows
that the \emph{shape} of the pipeline is preserved and that only the
\emph{token roles} rotate to their new positions:
\begin{enumerate}[leftmargin=1.6em,itemsep=0.15em,topsep=0.15em]
  \item The \textbf{Object} position now inherits the role previously
        played by the subject: Hidden AIE rises above $0.9$ in the
        early block, MLP AIE peaks around $0.9$ in layers $0$--$8$,
        and the Attention channel spikes to $\sim\!0.45$ at layer $0$,
        the same low-layer context-fusion signature reported for
        the subject side.
  \item The \texttt{[MASK\_subj]} position (mid-sentence) takes over
        in the top layers on \emph{all three} channels: Hidden and
        Attention $\mathrm{AIE}\!\approx\!0.5$ at the last layer, and
        MLP $\approx\!0.25$, mirroring the \texttt{[MASK]}-side
        readout of App.~\ref{app:mask-mirror}.
  \item Relation and Prefix positions remain flat throughout, i.e.\
        they still do not carry fact-specific causal mass, regardless
        of whether they precede or follow the masked span.
\end{enumerate}

This is precisely the property that legitimises the bidirectional access
state of Definition~\ref{def:access-state}. When we build a reverse key by
masking the subject and keeping the object visible, we are not constructing
an out-of-distribution probe; we are exercising the same pipeline in a
role-swapped configuration that the model already supports natively. It is
also, in our view, the mechanistic reason why \emph{Binary} / bidirectional
editing is a well-defined operation on a masked DLM but not on an
autoregressive one: only in the DLM does the subject position causally
depend on tokens that appear \emph{to its right}. Combined with the
cross-model replication of App.~\ref{app:dream-repro}, the pipeline used by
\methodname{} is thus doubly robust: it survives a change of \emph{model}
(LLaDA-8B $\leftrightarrow$ Dream-7B) and a change of \emph{mask layout}
(object-side $\leftrightarrow$ subject-side), which justifies treating the
forward and reverse keys as symmetric, first-class ingredients of a single
editing rule.

\section{Editor Implementation Details}
\label{app:hyperparams}

Table~\ref{tab:hyperparams} lists all hyperparameters shared across
LLaDA-8B and Dream-7B, and Table~\ref{tab:ctx-templates} enumerates the
context templates used for $z^*$ optimisation and key extraction.

\begin{table}[htbp]
\centering
\footnotesize
\caption{\textbf{Editing hyperparameters.}}
\label{tab:hyperparams}
\begin{tabular}{lll}
\toprule
\textbf{Category} & \textbf{Parameter} & \textbf{Value} \\
\midrule
\multirow{5}{*}{$z^*$ optimisation}
  & Optimiser         & Adam \\
  & Learning rate     & $5 \times 10^{-2}$ \\
  & Max steps         & 100 (early stop at loss $< 0.09$) \\
  & Weight decay $v_{\text{wd}}$ & $0.5$ \\
  & Clamp factor $\kappa$ & $3.0$ \quad ($\|\delta\| \le \kappa\|z_0\|$) \\
\midrule
\multirow{3}{*}{Weight update}
  & Cov.\ regularisation $\lambda$ & $10^{-5}$ \\
  & Base editor       & ROME (single layer) / MEMIT (multi-layer) \\
  & Write position    & last subject token (App.~\ref{app:position-sweep}) \\
\midrule
\multirow{3}{*}{FT+L baseline}
  & Learning rate     & $5 \times 10^{-4}$ \\
  & Steps             & 80 \\
  & \texttt{ft\_norm\_constraint} & $5 \times 10^{-4}$ (default) \\
\midrule
\multirow{5}{*}{Covariance $\hat{C}$}
  & Corpus $\mathcal{D}$ & WikiText-103 \\
  & Samples           & 100\,000 chunks ($\le 256$ tokens each) \\
  & Mask ratio        & $r \sim U(0.15,\,0.75)$ \\
  & Views / sentence  & 2 (independent $r$ and mask positions) \\
  & Key collection    & Non-masked positions only ($\sim$5M keys/layer) \\
\midrule
\multirow{4}{*}{Reproducibility and compute}
    & Sampling seed     & $42$ \\ 
  & Peak GPU memory   & $\approx 50{,}000$\,MiB per run \\
  & Precision         & bfloat16 backbone, fp32 for $z^*$ optimisation \\
\bottomrule
\end{tabular}
\end{table}

\begin{table}[htbp]
\centering
\footnotesize
\caption{\textbf{Context templates.} Each prompt is wrapped as
\texttt{[prefix] + prompt + [MASK] + [suffix]}. Suffix templates
exploit the DLM's bidirectional attention to condition on context
\emph{after} the prediction slot.}
\label{tab:ctx-templates}
\begin{tabular}{cl}
\toprule
\textbf{\#} & \textbf{Template (prefix \textbar\textbar{} suffix)} \\
\midrule
\multicolumn{2}{l}{\emph{Prefix templates (prepended before prompt):}} \\
1  & \texttt{""} \quad (identity, no wrapper) \\
2  & \texttt{"In one context, "} \\
3  & \texttt{"According to records, "} \\
4  & \texttt{"In this case, "} \\
5  & \texttt{"For reference, "} \\
6  & \texttt{"In general, "} \\
7  & \texttt{"As noted, "} \\
8  & \texttt{"At that time, "} \\
\midrule
\multicolumn{2}{l}{\emph{Suffix templates (appended after \texttt{[MASK]}):}} \\
9  & \texttt{", according to records"} \\
10 & \texttt{", as documented"} \\
11 & \texttt{", as indicated"} \\
12 & \texttt{", based on records"} \\
13 & \texttt{", in context"} \\
14 & \texttt{", in this instance"} \\
\bottomrule
\end{tabular}
\end{table}

\subsection{Covariance estimation under the masking distribution}
\label{app:update-details}

The update of \eqref{eq:rank1} is written along
$\hat{k} = C^{-1}k_{\mathrm{avg}}/(k_{\mathrm{avg}}^{\top}C^{-1}k_{\mathrm{avg}})$,
with
\begin{equation}
  C \;=\; \mathbb{E}_{\,x\sim\mathcal{D}}\;
  \mathbb{E}_{\,t\sim\mathcal{U}(0.15,0.75)}\;
  \mathbb{E}_{\,\tilde{x}\sim q_t(\cdot\mid x)}
  \Big[k^{(\ell)}(\tilde{x},\cdot)\,k^{(\ell)}(\tilde{x},\cdot)^{\top}\Big],
  \label{eq:cov}
\end{equation}
estimated offline on a corpus sample $\mathcal{D}$. Rescaling by the second
moment of keys the model actually produces suppresses directions that are
already high-variance across ordinary text, so that the edit does not fire on
unrelated inputs. The DLM-specific choice is the second expectation: keys are
collected from \emph{randomly masked} inputs, drawn from the same $q_t$ the
model traverses during decoding, not from clean text, so the statistics
mirror what the view family of \S\ref{sec:method:views} does in the
supervision. Concretely, we estimate the key covariance at each edit layer as
$\hat{C}_\ell = \mathbb{E}[\,k_\ell^{(i)}{k_\ell^{(i)}}^\top \mid
i \notin M\,]$, where the expectation marginalises over the DLM's
mask-ratio schedule and the conditioning $i \notin M$ restricts to
unmasked (content) positions, matching the distribution of keys
seen during actual editing. The inputs are masked; it is the
\emph{positions} at which keys are read that are restricted to unmasked
tokens. We invert with Tikhonov regularisation:
$\hat{C}_\ell^{-1} \leftarrow (\hat{C}_\ell + \lambda I)^{-1}$.
App.~\ref{app:ablation} isolates the choice: relative to clean-text
covariance, estimating the statistics under $q_t$ adds $2.6$~pp of paraphrase
transfer and improves neighborhood specificity and locality monotonically,
while reliability is unaffected. The covariance is estimated once per model
on WikiText-103 (Table~\ref{tab:hyperparams}) and cached.

\subsection{Reproducibility}
The editor involves no random initialisation or stochastic retraining.
Seed $42$ fixes sampled masking patterns and covariance estimation.
Each run fits in a single GPU with approximately
$50{,}000$\,MiB of memory, which is the peak observed for LLaDA-8B with the
largest view family; covariance estimation is performed once per model and
cached. We will release the full editing and evaluation code, together with the
cached covariance statistics and the view-construction scripts, upon acceptance.

\section{Editing Experiments: Details and Additional Results}
\label{app:editing-details}

\subsection{Benchmarks, scoring, and configuration}
\label{app:setup-details}

\paragraph{Models and benchmarks.}
CounterFact measures source reliability, paraphrase transfer, neighborhood
specificity and locality preservation; BAKE adds reverse QA and judgment
probes~\citep{meng2022rome,ma2023bird}; QAEdit supplies declarative edits and
scores freely generated answers to natural questions as exact ($1$), partial
($0.5$) or wrong ($0$)~\citep{yang2025mirage}. Neighborhood (Neigh.) scores
interference on \emph{similar} subjects. Locality preservation (Loc-Pres.) is
the fraction of unrelated prompts on which the edited model's top-ranked
continuation is unchanged from the pre-edit model, so it scores interference
on unrelated text.

\paragraph{Scoring.}
CounterFact and BAKE follow the ROME forced-choice criterion:
$o_{\mathrm{new}}$ must outrank $o_{\mathrm{true}}$ (or \texttt{Yes} must
outrank \texttt{No}). For QAEdit we report score-bucket frequencies and
$s=(n_{1.0}+0.5n_{0.5})/n$. We retain only examples the base model answers
fully wrongly ($\texttt{pre}<0.5$), so every non-zero score is attributable to
the edit.

\paragraph{Editor configuration.}
\textsc{Multi-View} changes key/value construction and can use a standard
weight-update rule underneath. Main results use ROME at layer $7$, written at
the last subject token; App.~\ref{app:position-sweep} reports the position
sweep. MEMIT spreads the same construction over layers $5$--$7$, all within
the mechanistic block. Concurrent DLM editors address a different failure
mode and install no object-side key, so they are not comparable on the
reverse probes of \S\ref{sec:bake} (App.~\ref{app:related-work}). On QAEdit,
where their correction does apply, we reproduce one of them and compose it
with ours (App.~\ref{app:compose-parkjo}). We include constrained fine-tuning
(FT+L) as a reproducible DLM baseline and show published AR results only as
scale references (App.~\ref{app:ar-refs}). Hyperparameters are listed in
App.~\ref{app:hyperparams}.

\subsection{Published autoregressive references}
\label{app:ar-refs}

Tables~\ref{tab:cf} and~\ref{tab:bake} compare \methodname{} against matched
single-view and single-direction variants and against constrained
fine-tuning on the same DLMs. Published results for autoregressive editors on
GPT-J (6B) and LLaMA-2 (7B), reproduced from the original
papers~\citep{meng2022rome,meng2023memit,ma2023bird}, are collected in
Table~\ref{tab:ar-refs} as scale references: they concern different models
and different edit pools, so they are not controlled comparisons. On the
reverse probes they show the pattern the main text refers to: Reverse-QA
gains of at most $2.9$ points from a forward edit. The AR Reverse-Judge deltas
should be read as upper bounds, because BAKE reports a pre-edit floor of zero
that is inherited from the QA probe rather than measured on the judgment
probe (see below).

\begin{table}[htbp]
\centering
\footnotesize
\setlength{\tabcolsep}{4pt}
\caption{\textbf{Published autoregressive references.} CounterFact and BAKE
results reported by the original papers for GPT-J (6B) and LLaMA-2 (7B).
BAKE reports a pre-edit score of $0.00$ on both reverse probes for every base
LLM, so the published $\Delta$ equals the post-edit value. BIRD does not
report CounterFact numbers.}
\label{tab:ar-refs}
\begin{tabular}{llccccccc}
\toprule
& & \multicolumn{3}{c}{\textbf{CounterFact}} & \multicolumn{4}{c}{\textbf{BAKE}} \\
\cmidrule(lr){3-5}\cmidrule(lr){6-9}
\textbf{Model} & \textbf{Method}
  & Rel.~\metup & Para.~\metup & Neigh.~\metup
  & Rel.~\metup & Para.~\metup & R-QA~\metup & R-Judge~\metup \\
\midrule
\multirow{3}{*}{GPT-J (6B)}
  & ROME  & 100.00 & 97.95 & 81.46 & 98.01 & 94.60 & 2.80 & 33.89 \\
  & MEMIT & 100.00 & 97.10 & 81.96 & 96.21 & 86.99 & 0.16 & 13.47 \\
  & BIRD  & --     & --    & --    & 99.33 & 96.00 & 2.90 & 34.96 \\
\midrule
\multirow{3}{*}{LLaMA-2 (7B)}
  & ROME  & 99.68 & 98.10 & 77.59 & 99.70 & 94.74 & 0.26 & 40.78 \\
  & MEMIT & 99.58 & 97.99 & 77.80 & 99.85 & 96.92 & 0.14 & 30.10 \\
  & BIRD  & --    & --    & --    & 99.90 & 96.20 & 0.28 & 42.72 \\
\bottomrule
\end{tabular}
\end{table}

\paragraph{The pre-edit floor of BAKE's judgment probe.}
\label{app:bake-floor}
BAKE reports a pre-edit score of zero on both reverse probes for every base
LLM, attributing this to the filtering of edits whose reverse fact the model
already holds. The filter, however, is defined only in the QA direction: given
$o^{*}$ and $r^{-1}$, an edit is retained if the base model prefers $s^{*}$
over $s$, which is exactly the RQS criterion $P(s) > P(s^{*})$. RJS scores a
different quantity, $P(\text{yes}) > P(\text{no})$ on a judgment prompt, and
nothing in the filter constrains it. A model that has been filtered to not
prefer $s$ in the QA direction may still answer \emph{yes} to ``Whether the
capital of England is Paris?'', through acquiescence bias, through
sensitivity to the surface form of the question, or simply because the
judgment prompt makes the target explicit rather than requiring it to be
generated. The zero is thus inherited from the QA probe rather than measured
on the judgment probe.

Two consequences. First, the AR $\Delta$ on R-Judge should be read as an upper
bound: whatever fraction of judgment prompts the unedited model already answers
affirmatively is being credited to the editor. Second, this asymmetry works
against the DLMs in a naive reading of Table~\ref{tab:bake}, since our
pre-edit rows are measured rather than assumed ($30.6\%$ on LLaDA-8B and
$18.2\%$ on Dream-7B) and are subtracted in full. That these measured floors
are far from zero on the same style of prompt is itself the clearest evidence
that a judgment probe carries a non-trivial affirmative baseline. 
Directly measuring the AR judgment floor on the released BAKE prompts would
settle the question and is a straightforward addition we did not have the
compute budget to include.

\subsection{Which subject position should the edit be written at?}
\label{app:position-sweep}

\paragraph{Why the question is different in a DLM.}
Locate-then-edit methods developed for AR models write the update at the
\emph{last} subject token. This convention is a consequence of the causal mask:
under left-to-right attention, an earlier subject token has not yet seen the
remainder of the subject span, and no subject token can see the relation
template placed to its right, so the last subject position is the only one whose
residual stream carries the complete $(s,r)$ context by the time the early MLPs
read it. The choice is therefore forced rather than empirical.

Section~\ref{sec:mechanism} shows that this argument does not carry over. Layer-$0$
attention in a masked DLM fuses the relation template into the subject residual
stream (Table~\ref{tab:l0-surgery}), and because attention is bidirectional it
does so at \emph{every} subject position simultaneously: each subject token
attends to the full relation span regardless of its index within the span. The
$(s,r)$-conditioned key of \eqref{eq:binding} is thus available at any
position inside $\mathcal{S}_s$, and which one we write the rank-one update at
becomes an empirical choice rather than a structural constraint. 

\paragraph{Selection protocol.}
Because the position is not fixed by the architecture and the design space is
small, we settle it empirically, but we separate selection from reporting. The
sweep is run on a development pool of $n{=}4694$ BAKE edits
(Table~\ref{tab:position-sweep}, top block), the winning setting is fixed, and
it is then re-evaluated once on a disjoint held-out split of $n{=}1000$ edits
that is used for no other tuning decision in this paper
(Table~\ref{tab:position-sweep}, bottom block). The held-out split reproduces
the development ranking exactly: \emph{last} is best on all four columns on both
splits, and the full three-way ordering is identical column by column
(\emph{last} $>$ \emph{mean} $>$ \emph{first} on Reliability and Paraphrase,
\emph{last} $>$ \emph{first} $>$ \emph{mean} on both reverse probes). The
Paraphrase advantage of \emph{last} shrinks from $17.1$ to $14.5$~pp under
held-out evaluation, i.e.\ some but not all of it was selection effect.

\begin{table}[htbp]
\centering
\footnotesize
\caption{\textbf{Subject-position sweep on LLaDA-8B (BAKE, ROME base editor).}
\emph{first} / \emph{mean} / \emph{last} write the update at the first subject
token, at the mean over the subject span, and at the last subject token
respectively. The top block is the development pool on which the position was
selected; the bottom block is a disjoint held-out split evaluated once after the
choice was fixed. Reliability and both reverse probes are insensitive to the
choice on both splits; Paraphrase is not. \emph{last} is best on all four
columns on both splits, and the three-way ordering is identical column by
column. The \emph{last} setting is the configuration reported throughout the
main text.}
\label{tab:position-sweep}
\begin{tabular}{llcccc}
\toprule
Split & Position & Rel.~\metup & Para.~\metup & R-QA~\metup & R-Judge~\metup \\
\midrule
\multirow{4}{*}{\shortstack[l]{Development\\($n{=}4694$)}}
 & first & 97.23 & 67.22 & 60.94 & 52.53 \\
 & mean  & 98.27 & 73.63 & 61.02 & 50.79 \\
 & last  & \best{98.49} & \best{84.56} & \best{63.24} & \best{55.60} \\
\cmidrule(l){2-6}
 & spread & 1.26 & \loss{17.34} & 2.30 & 4.81 \\
\midrule
\multirow{4}{*}{\shortstack[l]{Held-out\\($n{=}1000$)}}
 & first & 97.50 & 67.60 & 58.60 & 56.60 \\
 & mean  & 97.90 & 75.40 & 58.60 & 55.90 \\
 & last (main text) & \best{98.20} & \best{82.10} & \best{61.00} & \best{57.80} \\
\cmidrule(l){2-6}
 & spread & 0.70 & \loss{14.50} & 2.40 & 1.90 \\
\bottomrule
\end{tabular}
\end{table}

\paragraph{Findings.}
On the held-out split, three of the four columns are largely insensitive to
where in the subject span the update is written: Reliability varies by
$0.7$~pp, Reverse-QA by $2.4$~pp and Reverse-Judge by $1.9$~pp across the three
settings, differences that are small relative to the effect of the design
axes in \S\ref{sec:bake}. This is what \S\ref{sec:l0-surgery} predicts, and it
is the sense in which our result is \emph{not} a recovery of the AR convention:
that convention rests on the last subject token being the \emph{only} position
with a usable key, whereas here \emph{first} and \emph{mean} both remain
serviceable ($\ge 97.5\%$ Reliability, $\ge 58.6\%$ R-QA), i.e.\ every subject
position carries a usable $(s,r)$-conditioned key.

Paraphrase is the exception, ranging over $67.6\%$--$82.1\%$: a $14.5$~pp
spread on held-out data, comparable to the effect of the view family itself.
The write position therefore deserves treatment as a first-class design axis
rather than an implementation detail. We read the Paraphrase gap as a
quantitative effect layered on top of the layer-$0$ mechanism rather than a
structural one: positions later in the subject span aggregate more of the
surrounding context into the key even though all positions can access it, so an
update written there is triggered by a slightly broader region of key space.

\subsection{Ablations}
\label{app:ablation}
\paragraph{Matched view budget.}
\textsc{Multi-View} supervises the shared value over more views than
\textsc{Single-View} does, so its gains could in principle come from the extra
optimisation signal rather than from the relation-masked views themselves. We
therefore construct a budget-matched variant, \textsc{Multi-Binary} (matched),
that keeps the relation-masked and reverse construction but removes six views
from the context family, leaving it with the same number of views as
\textsc{Single-Binary}. Any remaining gap is then attributable to \emph{which}
views are used, not \emph{how many}. Tables~\ref{tab:ablation-budget-bake}
and~\ref{tab:ablation-budget-cf} report this comparison on BAKE and
CounterFact; all other components of the editor, including the base editor,
the write position, and the number of optimisation steps, are held fixed.

\begin{table}[htbp]
\centering
\footnotesize
\setlength{\tabcolsep}{4pt}
\caption{\textbf{Matched view budget on BAKE (LLaDA-8B).}
\emph{\#Views} counts the supervised views per direction.
\textsc{Multi-Binary} (matched) uses the relation-masked and reverse
construction with six context views removed, giving it the same budget as
\textsc{Single-Binary}. All other components are held fixed.}
\label{tab:ablation-budget-bake}
\begin{tabular}{lccccc}
\toprule
\textbf{Variant} & \#Views
  & Rel.~\metup & Para.~\metup & R-QA~\metup & R-Judge~\metup \\
\midrule
\textsc{Single-Binary}            & 14 & 97.04 & 65.48 & 42.03 & 41.03 \\
\textsc{Multi-Binary} (matched)   & 14 & 97.81 & 82.24 & 63.81 & 55.92 \\
\textsc{Multi-Binary}             & 20 & 98.49 & 84.56 & 63.24 & 55.60 \\
\bottomrule
\end{tabular}
\end{table}

\begin{table}[htbp]
\centering
\footnotesize
\setlength{\tabcolsep}{4pt}
\caption{\textbf{Matched view budget on CounterFact (LLaDA-8B).}
Same three configurations as Table~\ref{tab:ablation-budget-bake}, evaluated on
the CounterFact edit pool.}
\label{tab:ablation-budget-cf}
\begin{tabular}{lccccc}
\toprule
\textbf{Variant} & \#Views
  & Rel.~\metup & Para.~\metup & Neigh.~\metup & Loc-Pres.~\metup \\
\midrule
\textsc{Single-View}              & 14 & 96.87 & 53.65 & 57.13 & 68.31 \\
\textsc{Multi-View} (matched)     & 14 & 98.03 & 66.85 & 71.05 & 89.64 \\
\textsc{Multi-View}               & 20 & 98.14 & 67.92 & 73.73 & 89.92 \\
\bottomrule
\end{tabular}
\end{table}

\paragraph{Findings.}
At a matched budget of $14$ views, the relation-masked construction already
recovers almost all of the gain. On BAKE it lifts Paraphrase from $65.5\%$ to
$82.2\%$ and Reverse-QA from $42.0\%$ to $63.8\%$, against $84.6\%$ and
$63.2\%$ for the full $20$-view configuration; on Reverse-QA the matched
variant is in fact marginally ahead. CounterFact shows the same pattern, with
the matched variant within $1.1$~pp of the full configuration on Paraphrase
($66.9\%$ vs.\ $67.9\%$) and within $2.7$ and $0.3$~pp on Neighborhood and
Loc-Pres. The gains of \textsc{Multi-View} therefore come from \emph{which}
views the shared value is supervised on rather than from how many: adding six
further context views on top of the relation-masked family buys a further
$2.3$~pp of Paraphrase on BAKE and little elsewhere. Since the number of
installed directions is fixed by construction
(\S\ref{sec:method:objective}), this also rules out the alternative explanation
that the improvement merely reflects more values being written into the
weights.

\paragraph{KL locality regularisation.}
Table~\ref{tab:ablation-kl} moves the KL comparison out of the main results.
Overall, KL has little influence on the result: on LLaDA-8B all changes are
within $1.6$ percentage points, while on Dream-7B its modest locality gains
($0.9$--$1.9$ points) are offset by lower reliability and paraphrase accuracy.
We therefore use the simpler unregularised objective as the default.

\begin{table}[htbp]
\centering
\footnotesize
\setlength{\tabcolsep}{4.5pt}
\caption{\textbf{KL regularisation ablation on CounterFact.}
All rows use the ROME base editor.}
\label{tab:ablation-kl}
\begin{tabular}{llcccc}
\toprule
\textbf{Model} & \textbf{Variant}
  & Rel.~\metup & Para.~\metup & Neigh.~\metup & Loc-Pres.~\metup \\
\midrule
\multirow{2}{*}{Dream-7B}
  & \basecell{\textsc{Multi-View}}      & \basecell{99.88} & \basecell{99.24} & \basecell{73.04} & \basecell{88.29} \\
  & \textsc{Multi-View}$+$KL & 98.27 & 92.47 & 74.89 & 89.21 \\
\midrule
\multirow{2}{*}{LLaDA-8B}
  & \basecell{\textsc{Multi-View}}      & \basecell{98.14} & \basecell{67.92} & \basecell{73.73} & \basecell{89.92} \\
  & \textsc{Multi-View}$+$KL & 96.83 & 66.38 & 74.03 & 90.18 \\
\bottomrule
\end{tabular}
\end{table}

\paragraph{Whitening covariance.}
\eqref{eq:cov} estimates covariance from randomly masked inputs so that
whitening matches the DLM deployment distribution.
Table~\ref{tab:ablation-covariance} isolates that choice against clean-text
covariance and against no whitening at all, holding the model, base editor, data
split and every non-covariance hyperparameter fixed.

\begin{table}[htbp]
\centering
\footnotesize
\caption{\textbf{Whitening covariance ablation on CounterFact (LLaDA-8B).}
The model, base editor, data split, and all non-covariance hyperparameters are
held fixed.}
\label{tab:ablation-covariance}
\begin{tabular}{lcccc}
\toprule
\textbf{Whitening statistics}
  & Rel.~\metup & Para.~\metup & Neigh.~\metup & Loc-Pres.~\metup \\
\midrule
No whitening                    & 97.20 & 59.55 & 70.94 & 83.41 \\
Clean-text covariance           & 97.20 & 65.32 & 72.85 & 87.96 \\
Random-mask covariance (ours)   & \best{98.14} & \best{67.92} & \best{73.73} & \best{89.92} \\
\bottomrule
\end{tabular}
\end{table}

\paragraph{Findings.}
Reliability is essentially unaffected by the choice of statistics
($97.2\%$--$98.1\%$): whitening does not determine whether the association is
installed. It determines how well the installed direction generalises and how
much it leaks. Moving from no whitening to clean-text covariance recovers
$5.8$~pp of Paraphrase, and estimating the covariance under the model's own
masking distribution adds a further $2.6$~pp; Neighborhood and Loc-Pres.\
improve monotonically in the same order ($70.9\% \to 72.9\% \to 73.7\%$ and
$83.4\% \to 88.0\% \to 89.9\%$). The gap between the two whitened rows is the
part attributable to \eqref{eq:cov} specifically, and it is consistent with the
motivation given there: the keys an edit is actually queried from at decoding
time are drawn from $q_t$, not from clean text, so second-order statistics
collected under $q_t$ describe the deployment geometry more accurately.

\paragraph{Prefix and suffix contexts.}
Because a DLM conditions bidirectionally, contextual templates can be placed
before or after the masked prediction span. Table~\ref{tab:ablation-context}
compares three configurations on CounterFact: no contextual templates,
prefix-only templates, and the full prefix-plus-suffix family. All other
components of the editor are held fixed.

\begin{table}[htbp]
\centering
\footnotesize
\setlength{\tabcolsep}{4pt}
\caption{\textbf{Prefix/suffix context ablation on CounterFact.}
\emph{None} uses no contextual templates, \emph{Prefix} uses only templates
before the prompt, and \emph{Prefix\,$+$\,suffix} combines prefix and suffix
templates; the last row of each block is the \textsc{Multi-View} configuration
of Table~\ref{tab:cf}. All other components of the editor are held fixed.
Bold indicates the best result within each model and metric; note that
Neighborhood and Loc-Pres.\ are best without contextual views, i.e.\ the gains
in Rel.\ and Para.\ are traded against a small locality cost.}
\label{tab:ablation-context}
\begin{tabular}{llcccc}
\toprule
\textbf{Model} & \textbf{Context views}
  & Rel.~\metup & Para.~\metup & Neigh.~\metup & Loc-Pres.~\metup \\
\midrule
\multirow{3}{*}{LLaDA-8B}
  & None
  & 91.60 & 38.20 & \best{\textbf{76.08}} & \best{\textbf{92.48}} \\
  & Prefix
  & 95.40 & 49.40 & 75.10 & 91.90 \\
  & Prefix $+$ suffix
  & \best{\textbf{98.14}} & \best{\textbf{67.92}} & 73.73 & 89.92 \\
\midrule
\multirow{3}{*}{Dream-7B}
  & None
  & 84.72 & 68.20 & \best{\textbf{76.44}} & \best{\textbf{90.50}} \\
  & Prefix
  & 96.98 & 85.40 & 74.58 & 89.84 \\
  & Prefix $+$ suffix
  & \best{\textbf{99.88}} & \best{\textbf{99.24}} & 73.04 & 88.29 \\
\bottomrule
\end{tabular}
\end{table}

\paragraph{Findings.}
Contextual views substantially improve edit success and generalisation on both
models. Relative to using no context templates, prefix-plus-suffix supervision
raises Reliability by $6.5$ percentage points and Paraphrase by $29.7$ points
on LLaDA-8B; the corresponding gains on Dream-7B are $15.2$ and $31.0$ points.
Adding suffix templates on top of prefix-only supervision accounts for a
substantial share of the paraphrase gain on both models ($+18.5$ points on
LLaDA-8B and $+13.8$ on Dream-7B), confirming that the extra conditioning
direction available to a bidirectional model is not redundant with prefix
context. These improvements come with modest locality costs: compared with no
contextual templates, Neighborhood decreases by $2.4/3.4$ points and
Loc-Pres.\ by $2.6/2.2$ points on LLaDA/Dream. Overall, bidirectional
prefix-plus-suffix conditioning offers a favourable trade-off, producing large
gains in reliability and paraphrase transfer for a relatively small reduction
in locality.

\subsection{Stability under sequential editing}
\label{app:sequential-decay}

We additionally evaluate whether the edits remain stable when multiple facts are
installed sequentially. Starting from the same base model, we apply ROME updates
one at a time without restoring the original weights, and evaluate the
cumulatively edited model after up to $T=100$ edits. We compare
\textsc{Single-Binary} with \textsc{Multi-Binary} under the same editing
protocol.

\begin{figure}[htbp]
    \centering
    \safeincludegraphics[width=\linewidth]{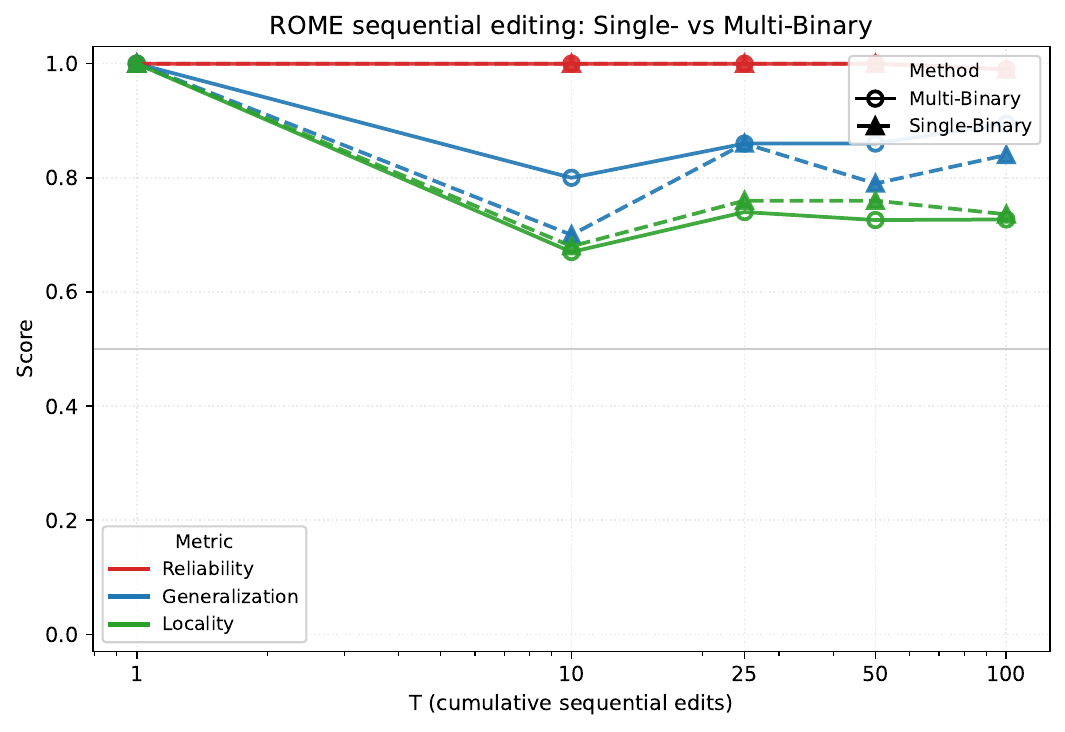}
    \caption{\textbf{ROME sequential-edit stability.}
    \textsc{Single-Binary} and \textsc{Multi-Binary} are applied cumulatively
    without restoring the model weights. Scores are evaluated after
    $T\in\{1,10,25,50,100\}$ sequential edits.}
    \label{fig:rome-sequential-decay}
\end{figure}

\paragraph{Findings.}
Reliability is unaffected by accumulation: both methods stay at $100\%$
through $T=25$ and at $\approx99\%$ after $100$ sequential edits, so
previously installed associations are not overwritten by later ones.
Generalisation and locality both take their largest hit between $T=1$ and
$T=10$ and then plateau or partly recover, and all three metrics remain well
above the $0.5$ reference line at every $T$. On generalisation the two
constructions separate: \textsc{Multi-Binary} drops to $80\%$ at $T=10$
against $70\%$ for \textsc{Single-Binary}, the two coincide at $86\%$ at
$T=25$, and thereafter \textsc{Multi-Binary} holds $86\%$ at $T=50$ and
$89.5\%$ at $T=100$ while \textsc{Single-Binary} falls to $79\%$ and $84\%$.
The wider key coverage therefore mainly buys resistance to generalisation
decay under accumulated interference: a $10$-point advantage at $T=10$ and
$5.5$ points at $T=100$. Locality is comparable between the two: from $100\%$
after a single edit it falls to about $67$--$68\%$ at $T=10$ for both methods
and settles at $76\%/74\%$ (Single/Multi) at $T=25$ and $73.6\%/72.7\%$ at
$T=100$, with \textsc{Multi-Binary} within $3.5$ points of
\textsc{Single-Binary} throughout. The stronger generalisation retention is
thus not bought with additional interference on unrelated examples.

\subsection{QAEdit: breakdown by target length}
\label{app:qaedit-length}

Table~\ref{tab:qaedit-len} breaks the LLaDA-8B source-prompt results of
Table~\ref{tab:qaedit} down by the token length of the gold answer.
Free-generation quality varies only mildly over $1$--$4$ word targets
($0.224$--$0.297$ weighted) and is lowest on $\ge\!5$-word targets ($0.198$),
i.e.\ the editor degrades gracefully rather than failing outright on
multi-token objects, but longer targets remain the harder case.
This is consistent with the multi-token degradation reported by
\citet{park2026editing}, whose correction we compose with our view family in
App.~\ref{app:compose-parkjo}.

\begin{table}[htbp]
\centering
\footnotesize
\setlength{\tabcolsep}{4pt}
\caption{\textbf{QAEdit LLaDA-8B breakdown by target length}
(source prompt, \textsc{Multi-View}).}
\label{tab:qaedit-len}
\begin{tabular}{lccccc}
\toprule
Target length & $n$
  & $=\!1.0$~\metup & $=\!0.5$ & $\ge\!0.5$~\metup & Weighted $s$~\metup \\
\midrule
1 word      & 677 & 21.3\% & 16.8\% & 38.1\% & \best{\textbf{0.297}} \\
2 words     & 820 & 16.2\% & 12.4\% & 28.7\% & 0.224 \\
3 words     & 437 & 16.2\% & 15.3\% & 31.6\% & 0.239 \\
4 words     & 176 & 18.2\% & 17.0\% & 35.2\% & 0.267 \\
$\ge\!5$    & 162 & 11.7\% & 16.0\% & 27.8\% & \loss{0.198} \\
\bottomrule
\end{tabular}
\end{table}

\subsection{Composing with decoding-state views: reproducing Park \& Jo on QAEdit}
\label{app:compose-parkjo}

App.~\ref{app:related-work} argues that the failure mode identified by
\citet{park2026editing} (a value fitted on a fully masked target is not
optimised for the partially revealed target states traversed during diffusion
decoding) is orthogonal to the key-coverage failure mode this paper
addresses, and that in the notation of \S\ref{sec:method:views} their
correction is simply a further view type: a partial \emph{reveal} of
$\mathcal{S}_o$ rather than an occlusion of $\mathcal{S}_r$. Because the two
concern different axes of the same editor, the additional views can be
supervised on the same shared value without any other change to
\eqref{eq:zstar}. This appendix tests that composition empirically.

QAEdit is the setting in which the comparison is well posed. The reverse probes
of \S\ref{sec:bake} cannot discriminate a forward-only editor at all, but QAEdit
scores freely decoded answers, which is exactly the regime the decoding-state
correction targets: every answer token beyond the first is produced from a
partially revealed target. We therefore reimplement the partial-reveal views on
LLaDA-8B and add them to $\mathcal{V}$, holding the view family, the key
aggregation \eqref{eq:kavg}, the whitening \eqref{eq:cov}, the write position
and every hyperparameter of Table~\ref{tab:hyperparams} fixed. The evaluation
pool is unchanged from \S\ref{sec:qaedit}.

\begin{table}[htbp]
\centering
\footnotesize
\setlength{\tabcolsep}{3.5pt}
\caption{\textbf{Composing decoding-state views with the multi-view family on
QAEdit} (LLaDA-8B, pre-edit-wrong examples). The \textsc{Multi-View} rows repeat
Table~\ref{tab:qaedit}. \textsc{$+$ partial reveal} adds the partially revealed
target states of \citet{park2026editing} to $\mathcal{V}$ as an additional view
type; all other components of the editor are held fixed.}
\label{tab:compose-parkjo}
\begin{tabular}{llccccc}
\toprule
\textbf{Method} & \textbf{Split}
  & $=\!1.0$~\metup & $=\!0.5$ & $=\!0.0$~\metdn & $\ge\!0.5$~\metup
  & Weighted $s$~\metup \\
\midrule
\multirow{2}{*}{\textsc{Multi-View}}
  & Rel. & 17.6\% & 14.9\% & 67.5\% & 32.5\% & 0.250 \\
  & Gen. & 12.6\% & 12.9\% & 74.5\% & 25.5\% & 0.187 \\
\cmidrule(lr){2-7}
\multirow{2}{*}{\textsc{Multi-View} $+$ partial reveal}
  & Rel. & \best{\textbf{35.0\%}} & 16.1\%
        & \best{\textbf{48.9\%}} & \best{\textbf{51.1\%}}
        & \best{\textbf{0.430}} \\
  & Gen. & \best{\textbf{26.7\%}} & 17.0\%
        & \best{\textbf{56.3\%}} & \best{\textbf{43.7\%}}
        & \best{\textbf{0.352}} \\
\bottomrule
\end{tabular}
\end{table}

\begin{table}[htbp]
\centering
\footnotesize
\setlength{\tabcolsep}{4pt}
\caption{\textbf{QAEdit LLaDA-8B breakdown by target length for the composed
editor} (source prompt, \textsc{Multi-View} $+$ partial reveal).
Table~\ref{tab:qaedit-len} reports the same breakdown for \textsc{Multi-View}
alone.}
\label{tab:compose-len}
\begin{tabular}{lccccc}
\toprule
Target length 
  & $=\!1.0$~\metup & $=\!0.5$ & $\ge\!0.5$~\metup & Weighted $s$~\metup \\
\midrule
1 word     & 34.3\% & 18.2\% & 52.5\% & \best{\textbf{0.434}} \\
2 words      & 25.9\% & 12.0\% & 38.0\% & 0.319 \\
3 words      & 21.3\% & 21.3\% & 42.7\% & 0.320 \\
4 words      & 21.7\% & 17.4\% & 39.1\% & 0.304 \\
$\ge\!5$     & 20.0\% & 20.0\% & 40.0\% & 0.300 \\
\bottomrule
\end{tabular}
\end{table}

\paragraph{Findings.}
The two corrections compose. On source prompts the weighted score rises from
$0.250$ to $0.430$ and the fraction of previously wrong answers made at least
partially correct from $32.5\%$ to $51.1\%$, so the composed editor is at least
partly right on a majority of the examples the unedited model got fully wrong;
on paraphrased questions the corresponding figures are $0.187 \to 0.352$ and
$25.5\% \to 43.7\%$. Fully correct answers roughly double on both splits. The
gain is obtained \emph{on top of} the full view family of
\S\ref{sec:method:views}, with the key construction and the number of installed
directions unchanged, so it is attributable to the view type
\citet{park2026editing} contribute rather than to additional optimisation signal
along an axis we already cover.

Table~\ref{tab:compose-len} locates that gain where the two-failure-mode reading
predicts it. The multi-token degradation of Table~\ref{tab:qaedit-len} is
largely removed: the $\ge\!5$-word bucket, the worst case for
\textsc{Multi-View} at $0.198$, reaches $0.300$, and every multi-token bucket
now scores above the best single-token result of the uncomposed editor
($0.297$). What remains is a single step from one-word to multi-word targets
($0.434$ vs.\ $0.300$--$0.320$) rather than continued decay with length: across
the $2$--$\ge\!5$ buckets the spread falls from $6.9$ to $2.0$ points. The
component of the length effect that the decoding-state views remove is thus the
component that accumulates with the number of revealed target tokens, as their
diagnosis implies; the residual one-word advantage is not a decoding-trajectory
effect and we do not account for it here.

\section{A Key-Space Account of Statement-Level Generalisation}
\label{app:theory}

The design principle of \S\ref{sec:method:principle} (supervise the edit on
views that span the region of key space the fact is queried from) can be
made precise. We define generalisation as an objective over the conditionals a
fact induces at deployment, note that under the rank-one update of
\eqref{eq:rank1} this objective is governed entirely by the trigger
strength of \eqref{eq:alpha}, and show that (i)~the expected trigger over
a query family is maximised by the family's mean key, with the single-view
editor suffering a closed-form attenuation (Prop.~\ref{prop:coverage});
(ii)~no single rank-one association serves both query directions, one per
direction suffices, and reverse generalisation decomposes into a direction
term and a within-direction coverage term (Prop.~\ref{prop:direction});
(iii)~the finite view family of \S\ref{sec:method:views} attains the
population optimum up to a view-shift term controlled by sampling from the
model's own $q_t$ and a sampling term decaying with the view budget
(Prop.~\ref{prop:finite}). The treatment is a first-order analysis of the
editor's linear algebra; its one modelling assumption is stated explicitly and
is checkable against the measured triggers of \S\ref{sec:multiview-analysis}.

\subsection{Setup and notation}
\label{app:theory-setup}

\paragraph{Masked-diffusion objective.}
A masked DLM $p_\theta$ is trained to reconstruct randomly hidden tokens.
Given $x$ and a mask rate $t\!\sim\!\mathcal{U}(0,1)$, the corruption operator
$q_t$ replaces each token independently by \Mask{} with probability $t$, and
the training loss is the (negative) ELBO
\begin{equation}
  \mathcal{L}_{\mathrm{ELBO}}(x)
  \;=\;
  \mathbb{E}_{t\sim\mathcal{U}(0,1)}\;
  \mathbb{E}_{\tilde{x}\sim q_t(\cdot\mid x)}
  \Big[\tfrac{1}{t}\!\!\sum_{p\,:\,\tilde{x}_p=\Mask}\!\!
  -\log p_\theta\!\left(x_p \mid \tilde{x}\right)\Big].
  \label{eq:elbo}
\end{equation}
Two features of \eqref{eq:elbo} underlie the analysis below. Where an AR
model is trained on $p(o\mid s,r)$ alone, \eqref{eq:elbo} trains the model
on every conditional in which one subset of $x$ is visible and another is
predicted, so the forward and reverse queries of Definition~\ref{def:Q} are
draws from the same family the model was trained on. And because the model is
queried under $q_t$ at decoding time as well, the keys an edit meets at test
time come from partially masked inputs; this is why both the view family of
\S\ref{sec:method:views} and the whitening covariance of \eqref{eq:cov}
are drawn under $q_t$.

\paragraph{Editor notation.}
Fix an edit $(s,r,o)\!\to\!(s,r,o^\ast)$ and an edit layer (suppressed
throughout; the MEMIT spreading of \eqref{eq:rank1} divides one residual
across layers and changes nothing below). Let $C\succ0$ be the regularised
second moment of \eqref{eq:cov}, estimated on keys of the corpus
$\mathcal{D}$ under the masking distribution $q_t$; we write $\tilde{D}$ for
that key distribution and $\varphi(k)=C^{-1/2}k$ for the whitened key, so that
$\mathbb{E}_{\tilde D}[\varphi(k)\varphi(k)^\top]=I$ up to the Tikhonov term.
Let $a$ denote the aggregation key the update is written along
($a=k_{\mathrm{avg}}$ in \eqref{eq:kavg}). In whitened coordinates
\eqref{eq:alpha} reads
\begin{equation}
\alpha_a(k)=\frac{\langle\varphi(a),\varphi(k)\rangle}{\lVert\varphi(a)\rVert^2},
\qquad \alpha_a(a)=1 .
\label{eq:trigger-whitened}
\end{equation}

\begin{definition}[Fact-query distribution]
\label{def:Q}
At deployment the fact is interrogated through masked conditionals under the
corruption $q_t$ of \eqref{eq:elbo}: object-prediction queries with
varying relation wording, context and mask pattern (forward, $Q_f$), and
subject-prediction queries from the object side (reverse, $Q_r$). Model
this as the mixture $Q=\pi_fQ_f+\pi_rQ_r$ with weights $\pi_f+\pi_r=1$. For a
query $q$ let $\kappa(q)$ be the key read at its retrieval position (a subject
position forward, the first token of $o^\ast$ in reverse; \S\ref{sec:method:update})
and $\varphi_q=\varphi(\kappa(q))$. For $d\in\{f,r\}$ write
$\mu_d=\mathbb{E}_{Q_d}[\varphi_q]$ and
$\Sigma_d=\operatorname{Cov}_{Q_d}(\varphi_q)$, and assume $\mu_d\neq0$.
\end{definition}

By \eqref{eq:binding} the key depends on the relation's surface form and
on direction, so $\Sigma_d\neq0$ and $\mu_f\neq\mu_r$: the queries about one
fact occupy a region of key space, not a point. $\mu_d\neq0$ says the region
has a common component, which is necessary for any rank-one edit to cover it.

\begin{assumption}[Linear response]
\label{ass:linear}
Fix the installed value $z$. For a query $q$ in direction $d$, the editing
effect $\Delta\mathrm{logit}(q)$ of \S\ref{sec:multiview-analysis} (with the
target subject in place of $o_{\mathrm{new}}$ for reverse queries) satisfies
$\Delta\mathrm{logit}(q)=\gamma_d\,\alpha_a(\kappa(q))$, where $\gamma_d>0$ if
$z$ is \emph{adequate} on $Q_d$ (attains the stopping loss of
\eqref{eq:zstar} on that family) and $\gamma_d\le0$ otherwise. A query
succeeds iff $\alpha_a(\kappa(q))\ge\tau_q$ for some threshold $\tau_q<1$.
\end{assumption}

This is the single step bridging the editor's linear algebra to scores: it
takes the additive form of a rank-one update literally and ignores second-order
interaction with the frozen network. It is directly falsifiable, because
\S\ref{sec:multiview-analysis} measures $\alpha_{\mathrm{fire}}$ and
$\Delta\mathrm{logit}$ per edit and per query type. It predicts
(a)~$\Delta\mathrm{logit}\approx0$ wherever $\alpha\approx0$, as observed for
M1 on Reverse-QA ($\alpha=0.084$, $\Delta\mathrm{logit}=-0.028$); (b)~at a
fixed value, equal $\alpha$ implies equal $\Delta\mathrm{logit}$: M1 and M2
share the forward value and give $(0.609,1.500)$ vs.\ $(0.609,1.511)$ on
paraphrase; (c)~a per-edit regression of $\Delta\mathrm{logit}$ on
$\alpha_{\mathrm{fire}}$ within a query type should be approximately linear
through the origin, its slope estimating $\gamma_d$. Where $\alpha$ is constant
but $\Delta\mathrm{logit}$ moves (M1$\to$M3 on paraphrase, $0.609\to0.624$
vs.\ $1.500\to3.307$), the assumption attributes the change to $\gamma_d$,
i.e.\ to the value: the family supervision of \eqref{eq:zstar} improves
value adequacy on paraphrases, a channel separate from key coverage. For an
edit with aggregation keys $a_f$ and $a_r$, let $d(q)$ select the direction
of query $q$. Under Assumption~\ref{ass:linear}, the expected activation over
the fact-query distribution is
\begin{equation}
G(a_f,a_r)=\mathbb{E}_{q\sim Q}\big[\alpha_{a_{d(q)}}(\kappa(q))\big]
=\pi_f\,G_f(a_f)+\pi_r\,G_r(a_r),
\qquad
G_d(a)=\mathbb{E}_{Q_d}\big[\alpha_a(\kappa(q))\big],
\label{eq:objective}
\end{equation}
with $G_d$ its per-direction surrogate (dependence on the value suppressed).
Under Assumption~\ref{ass:linear} success on $q$ is a threshold on $\alpha$, so
maximising \eqref{eq:objective} maximises expected accuracy to first order.
\subsection{Coverage: the view mean maximises the expected trigger}
\label{app:theory-coverage}

\begin{proposition}[Expected trigger and single-view attenuation]
\label{prop:coverage}
Fix a direction, write $\mu=\mu_d,\Sigma=\Sigma_d$, and decompose
$\varphi(a)=\mu+e$ for an arbitrary error vector $e$. Then
\begin{equation}
G_d(a)=
\frac{\lVert\mu\rVert^2+\langle e,\mu\rangle}
     {\lVert\mu\rVert^2+2\langle e,\mu\rangle+\lVert e\rVert^2}.
\label{eq:exact-trigger}
\end{equation}
\begin{enumerate}
\item[(i)] \emph{(Population multi-view.)} If $\varphi(a)=\mu$ then
$G_d(a)=1$, $\operatorname{Var}_{Q_d}[\alpha_a]=\mu^\top\Sigma\mu/\lVert\mu\rVert^4$,
and for any $\tau<1$
\begin{equation}
\mathbb{P}_{Q_d}[\alpha_a\ge\tau]\;\ge\;
1-\frac{\mu^\top\Sigma\mu}{\mu^\top\Sigma\mu+\lVert\mu\rVert^4(1-\tau)^2}.
\end{equation}
\item[(ii)] \emph{(Single view.)} If $\varphi(a)=\mu+\nu$ with
$\langle\nu,\mu\rangle=0$, then
$G_d(a)=1/(1+\beta^2)$ with dispersion ratio $\beta=\lVert\nu\rVert/\lVert\mu\rVert$.
\end{enumerate}
\end{proposition}

\begin{proof}
$\mathbb{E}_{Q_d}[\alpha_a]=\langle\varphi(a),\mu\rangle/\lVert\varphi(a)\rVert^2$;
substituting $\varphi(a)=\mu+e$ gives \eqref{eq:exact-trigger}. For (i),
$\alpha_a-1=\langle\mu,\varphi_q-\mu\rangle/\lVert\mu\rVert^2$ has the stated
variance and Cantelli's inequality gives the tail bound. (ii) sets
$\langle e,\mu\rangle=0$ in \eqref{eq:exact-trigger}.
\end{proof}

\begin{remark}
\label{rem:singleview}
\textsc{Single-View} fits $a$ to the key of one edit sentence; its deviation
$\nu$ from $\mu_d$ is exactly the surface-form and residual-context component
that \eqref{eq:binding} identifies and that $\Vrel$ and $\Vctx$
marginalise out. Prop.~\ref{prop:coverage} thus locates the paraphrase failure
of prompt-fitted editors in one scalar, $\beta^2$, and predicts that replacing
the point key by the family mean removes the attenuation without increasing
the installed rank; (ii) is checkable as
$\hat\beta^2=1/\bar\alpha_{\mathrm{M1}}-1$, where $\bar\alpha_{\mathrm{M1}}$ is
the mean single-view trigger in Fig.~\ref{fig:multiview}. Maximising $G_d$
alone would be degenerate (shrinking $\lVert a\rVert$ inflates $\alpha_a$ on
every query), but because $\mathbb{E}_{\tilde D}[\varphi\varphi^\top]=I$, the
signal-to-leakage ratio
$(\mathbb{E}_{Q_d}\alpha_a)^2/\mathbb{E}_{\tilde D}[\alpha_a^2]
=\langle\varphi(a),\mu_d\rangle^2/\lVert\varphi(a)\rVert^2$ is scale-free and
maximised by the same direction $\mu_d$ (Cauchy--Schwarz). Mean aggregation is
therefore also specificity-optimal, consistent with the unchanged neighborhood
trigger of \S\ref{sec:multiview-analysis}; this identity is what
deployment-matched whitening (\eqref{eq:cov}) buys.
\end{remark}

\subsection{Direction: reversal requires, and only requires, a second association}
\label{app:theory-direction}

\begin{assumption}[$\varepsilon$-separated directions]
\label{ass:sep}
$|\langle\mu_f,\mu_r\rangle|\le\varepsilon\lVert\mu_f\rVert\lVert\mu_r\rVert$
and $\lVert\mu_r\rVert/\lVert\mu_f\rVert,\ \lVert\mu_f\rVert/\lVert\mu_r\rVert\le R$
for constants $\varepsilon\in[0,1)$, $R\ge1$. Moreover no single value is
adequate on both $Q_f$ and $Q_r$ (they emit different tokens at different
roles).
\end{assumption}

Forward and reverse keys are read at different token roles after the
role-swapped fusion of App.~\ref{app:mask-subject}; the measured trigger of a
forward-only edit on reverse queries, $\alpha\approx0.08$ (M1,
Fig.~\ref{fig:multiview}), gives $\varepsilon R\lesssim0.1$ empirically.

\begin{proposition}[The direction axis]
\label{prop:direction}
Under Assumptions~\ref{ass:linear}--\ref{ass:sep}:
\begin{enumerate}
\item[(i)] If $\varphi(a)=\mu_f$ then, whatever the value,
$G_r(a)=\langle\mu_f,\mu_r\rangle/\lVert\mu_f\rVert^2\le\varepsilon R$:
forward-only editing leaves reverse success within $O(\varepsilon R)$ of the
unedited baseline, so architectural availability of the reverse conditional
does not confer reachability.
\item[(ii)] For any single pair $(a,z)$ the value is inadequate on at least one
of $Q_f,Q_r$, where $\gamma_d\le0$: a rank-one edit improves at most one
direction.
\item[(iii)] Installing $\{(a_f,z_f^\ast),(a_r,z_r^\ast)\}$ with
$\varphi(a_d)=\mu_d$ and each $z_d^\ast$ adequate on $Q_d$
(\S\ref{sec:method:objective}--\ref{sec:method:update}) gives within-direction
expected trigger $1$ in each family, cross-direction interference at most
$\varepsilon R$ in expectation, and hence
$G\ge\pi_f\mathbb{P}_{Q_f}[\alpha_{a_f}\!\ge\tau]+
\pi_r\mathbb{P}_{Q_r}[\alpha_{a_r}\!\ge\tau]-O(\varepsilon R)$, with each
probability bounded as in Prop.~\ref{prop:coverage}(i) and the installed rank
fixed at one per direction independently of $|\mathcal{V}|$.
\end{enumerate}
\end{proposition}

\begin{proof}
(i) is $\mathbb{E}_{Q_r}[\alpha_a]=\langle\mu_f,\mu_r\rangle/\lVert\mu_f\rVert^2$
plus Assumption~\ref{ass:sep}. (ii): on the inadequate family
$\Delta\mathrm{logit}=\gamma_d\alpha\le0$. (iii): let
$\Delta_d=(z_d^\ast-\Wdown a_d)/m$ be the residual installed for direction $d$
in \eqref{eq:rank1}. A direction-$d$ query with key $\kappa$ receives
$\Delta_d\,\alpha_{a_d}(\kappa)+\Delta_{d'}\,\alpha_{a_{d'}}(\kappa)$, $d'\neq d$;
the second coefficient has expectation
$\langle\mu_{d'},\mu_d\rangle/\lVert\mu_{d'}\rVert^2\le\varepsilon R$ by
Assumption~\ref{ass:sep}, and the first reduces to Prop.~\ref{prop:coverage}(i).
\end{proof}

\begin{remark}[Reversal is inversion plus rewording]
\label{rem:reversal}
Reverse probes realise $r^{-1}$ in fresh wording, so $Q_r$ has its own
dispersion $\beta_r>0$, and the single reverse-view key of
\textsc{Single-Binary} deviates from $\mu_r$ by an inversion-template and a
wording component. Applying Prop.~\ref{prop:coverage} within $Q_r$:
\textsc{Single-Forward} attains $G_r\le\varepsilon R$
(Prop.~\ref{prop:direction}(i)); \textsc{Single-Binary} lifts this to
$1/(1+\beta_r^2)$; the reverse occlusions $\mathcal{R}_\rho$ of
\eqref{eq:revview} marginalise the wording component and move $G_r$
toward $1$ (\textsc{Multi-Binary}). Reverse generalisation decomposes exactly
into a direction term and a within-direction coverage term (the two axes of
\S\ref{sec:method:views}), predicting the two-step gains of
Table~\ref{tab:bake} ($22.4\%\to42.0\%\to63.2\%$ R-QA on LLaDA-8B) and their
persistence at a matched view budget (Table~\ref{tab:ablation-budget-bake}).
\end{remark}

\subsection{Finite views, and why they are drawn from $q_t$}
\label{app:theory-finite}

Let $P_V$ denote the distribution over direction-$d$ views induced by the
construction of \S\ref{sec:method:views} (occlusion ratio drawn from
$\mathcal{P}_\rho$, occlusion pattern, context template), and write
$\mu_{P_V}=\mathbb{E}_{v\sim P_V}[\varphi(k(v))]$ and $\Sigma_V$ for the mean and
covariance of the whitened view keys.

\begin{proposition}[Approximate maximisation from finitely many views]
\label{prop:finite}
Let $v_1,\dots,v_n\overset{\mathrm{iid}}{\sim}P_V$ and let $\hat a$ be their
aggregated key (\eqref{eq:kavg}; $\varphi$ is linear, so
$\varphi(\hat a)$ is the mean of the whitened view keys). For a failure
probability $\delta\in(0,1)$ put
$\eta=\big(\lVert\mu_{P_V}-\mu_d\rVert+\sqrt{\operatorname{tr}\Sigma_V/(n\delta)}\big)/\lVert\mu_d\rVert$.
With probability $\ge1-\delta$ over the draw of the views,
\begin{equation}
G_d(\hat a)\;\ge\;\frac{1-\eta}{(1+\eta)^2}\;\ge\;1-3\eta .
\end{equation}
\end{proposition}

\begin{proof}
$\varphi(\hat a)=\mu_d+e$ with $e=(\mu_{P_V}-\mu_d)+\xi_n$, where $\xi_n$ is
the sampling deviation of the empirical mean,
$\mathbb{E}\lVert\xi_n\rVert^2=\operatorname{tr}\Sigma_V/n$. Chebyshev gives
$\lVert e\rVert\le\eta\lVert\mu_d\rVert$ w.p.\ $\ge1-\delta$, and
Cauchy--Schwarz in \eqref{eq:exact-trigger} yields the first bound;
$(1-\eta)-(1-3\eta)(1+\eta)^2=5\eta^2+3\eta^3\ge0$ gives the second.
\end{proof}

The bound separates two errors. The sampling term decays with the view budget
$n$. The view-shift term $\lVert\mu_{P_V}-\mu_d\rVert$ does not decay with
$n$: it depends only on which distribution the views are drawn from.
Generating views with the model's own $q_t$ (the operator that defines the
deployment conditionals in \eqref{eq:elbo}) and marginalising the
occlusion ratio over $\mathcal{P}_\rho$ ties $P_V$ to $Q_d$ structurally,
without a paraphrase resource; \textsc{Single-View} is the degenerate case in
which $P_V$ is a point mass at the edit sentence, whose shift equals the full
dispersion $\beta\lVert\mu\rVert$ of Prop.~\ref{prop:coverage}(ii) and which no
budget reduces. This is the reading of the budget-matched ablation
(Tables~\ref{tab:ablation-budget-bake}--\ref{tab:ablation-budget-cf}): at
fixed $n$, changing the composition of the family changes $\mu_{P_V}$, and
nearly all of the gain moves with it.

\subsection{Predictions and scope}
\label{app:theory-predictions}

The account predicts, and the paper confirms: (P1)~forward-only editing leaves
reverse metrics at baseline despite the reverse conditional being available
(Prop.~\ref{prop:direction}(i); \textsc{Single-Forward}, Table~\ref{tab:bake});
(P2)~reverse gains arrive in two separable steps (Remark~\ref{rem:reversal};
Table~\ref{tab:bake}); (P3)~at matched budget, gains track the composition of
the view family, not its size (Prop.~\ref{prop:finite};
Tables~\ref{tab:ablation-budget-bake}--\ref{tab:ablation-budget-cf});
(P4)~widening the family leaves the unrelated-query trigger unchanged
(Remark~\ref{rem:singleview}; $\alpha\approx0.03$ in
\S\ref{sec:multiview-analysis}); (P5)~paraphrase gains split into a small
trigger and a large value component (Assumption~\ref{ass:linear}(c);
Fig.~\ref{fig:multiview}).

The analysis is idealised: Assumption~\ref{ass:linear} linearises a small
rank-one update; $\varepsilon,R,\beta$ and the view shift are measured, not
derived from the architecture; the treatment is per-fact and does not model
sequential interference (App.~\ref{app:sequential-decay}) or the
decoding-trajectory failure mode, which is orthogonal to the key-coverage axis
(App.~\ref{app:related-work}, App.~\ref{app:compose-parkjo}). Within these
limits, the view family is not an augmentation heuristic but a Monte-Carlo
estimator of the two population directions, one per query direction, that
maximise \eqref{eq:objective} under the editor's own update rule.

\section{Evidence Retrieval: Benchmarks, Protocols and Controls}
\label{app:retrieval}

\subsection{Controlled benchmark from PopQA facts}
\label{app:retrieval-popqa}

\paragraph{Corpus and queries.}
The controlled benchmark is built from PopQA facts~\citep{mallen2023popqa}
and is strictly balanced over five domains (Table~\ref{tab:retrieval-domains}).
It contains $1{,}382$ entities and $5{,}528$ evidence cards over $19$
distinct relations (citizenship appears in both the person and the sports
domain); every entity carries exactly the four relations of its domain. Each
card states the subject first, then its relation and the source answer
string(s), for example \emph{For ``Film A'', the director is Person X.}
Ambiguous names are removed globally and the entity sets of different
domains are disjoint. Each domain contributes $20$ validation and $60$ test
entities, giving $3{,}600$ validation and $10{,}800$ test queries; each test
domain has $720$ single-target, $720$ adjacent dual-target and $720$ crossed
dual-target queries, so the five domains and three layouts are weighted
equally. No example is filtered by whether a model answers it. Adjacent
queries place each entity next to its relation clause (\emph{For ``A'',
identify the director; for ``B'', identify the producer}); crossed queries
name both entities first and assign relations by reference (\emph{For ``A''
and ``B'', identify the director for the former and the producer for the
latter}). Entity spans are recorded when rendering each template.

\begin{table}[htbp]
\centering
\footnotesize
\caption{\textbf{Domains of the controlled benchmark.}}
\label{tab:retrieval-domains}
\begin{tabular}{llr}
\toprule
\textbf{Domain} & \textbf{Relations} & \textbf{Entities} \\
\midrule
Film production   & director, composer, screenwriter, producer     & 590 \\
Person attributes & occupation, citizenship, birthplace, language  & 351 \\
Sports people     & sport, team, playing position, citizenship     & 236 \\
Family relations  & father, mother, spouse, sibling                & 116 \\
Written works     & author, genre, publisher, language             &  89 \\
\bottomrule
\end{tabular}
\end{table}

\paragraph{Readouts and baselines.}
For subject retrieval, the vector for target $j$ is the hidden state at the
last subject token (L0, plain input) or the mean over its subject span at
the selected deeper layer under the retrieval prompt; each card is
represented at its own subject span, and the branch-specific score is the
cosine similarity between the two. Whole-query pooling averages the hidden
states of the original input text, excluding chat wrappers, padding and
representative-token suffixes. DiffRetriever uses its released zero-shot
single-step encoding with four query and sixteen passage representations;
PromptReps uses up to twenty sequential representative tokens on
Qwen2.5-7B-Instruct. Dense scoring uses normalized vectors with nonnegative
MaxSim; hybrid scoring equally combines min--max-normalized dense and sparse
scores. BM25 uses Porter stemming. One setting per method maximizes the
validation mean of single R@1, adjacent All@2 and crossed All@2, and all
settings are fixed before test evaluation.

\paragraph{Protocols.}
In \emph{global search} each subject branch ranks the full corpus; retrieval
alternates between branches and removes duplicate cards until the total
budget of two is met, with no entity-based candidate filter. Global pooling
and the official retrievers return a single ranking. Single R@1 measures
retrieval of the exact target card; dual All@2 measures whether both target
cards occur within two retrieved cards. The \emph{entity-conditioned}
protocol gives every method the same visible entity spans and an exact
unique-title linker: each target has four candidate cards (its four
relations), and at budget two each entity receives one card, so the protocol
isolates relation choice from entity coverage. Global methods reuse their
scores within each candidate set; hybrid normalization and fusion are
performed within it. Readout selection is repeated on validation under this
protocol.

\paragraph{Findings.}
On two-entity queries every single-vector method collapses in global
search: the best adjacent and crossed All@2 among BM25, pooling, DiffRetriever
and PromptReps are $3.06$ and $2.50$. Per-entity access states retrieve both
cards for $38.28$--$41.94\%$ of adjacent and $21.78$--$38.08\%$ of crossed
queries from the deeper layers, and for $13.32$--$24.44\%$ and
$9.44$--$13.33\%$ from the first block. The entity-conditioned protocol,
which removes entity coverage as a factor, leaves the ordering unchanged:
subject L31 reaches $41.94/38.08$ against at most $16.35/16.26$ for the
baselines, so the advantage lies in relation choice within an entity's cards,
not in finding the entity. Single-entity search over the full corpus is where
the representative-token and lexical retrievers are strongest (DiffRetriever hybrid
$97.69$, BM25 $92.82$, against $89.08$ for subject L31); the deeper readouts
also lead the first-block readout there, in contrast to the natural setting
of App.~\ref{app:retrieval-natural}, where the candidate set is a question's
own context and the first-block state is the best allocator.

\subsection{Natural comparison-question RAG}
\label{app:retrieval-natural}

We evaluate multi-entity evidence retrieval and answer generation on the
comparison subsets of HotpotQA and 2WikiMultiHopQA. A question is retained
when its official ten-paragraph distractor context provides two unambiguous
entity mentions and two valid gold evidence groups.
We reserve 100 questions from each dataset for validation and evaluate on
all remaining eligible questions: 1,296 from HotpotQA and 2,933 from
2WikiMultiHopQA. One additional 2Wiki question is excluded because its
annotated supporting-sentence index is outside the corresponding paragraph.
The resulting frozen corpus contains 106,329 unique Wikipedia sentences, each
prefixed with its paragraph title. For every question, a retriever ranks the
sentences in its official distractor context and returns two total
sentences. These sentences are placed in an identical prompt for a frozen
Qwen2.5-7B-Instruct reader, which generates a short answer with greedy
decoding.

We compare BM25, whole-query pooling, PromptReps, the released dense and
hybrid DiffRetriever variants, and subject-state retrieval on Dream-7B and
LLaDA-8B. Dream and LLaDA DiffRetriever use four query and sixteen passage
representations; Qwen PromptReps uses twenty of each. L0 subject retrieval
reads the last subject token from the plain input. The deeper subject
variants read the mean subject state at Dream L27 or LLaDA L31 under the
retrieval prompt. For a two-entity question, subject retrieval forms one
ranking per entity, alternates between the two rankings, and removes
duplicate sentences until the shared budget of two is filled. All
configurations are fixed before test evaluation. Support R@2 measures recall
over annotated supporting sentences, while Both@2 measures whether the
retrieved pair covers both gold evidence titles. End-to-end answer quality is
reported with normalized exact match and token F1. Latency is the median
per-query time from question to retrieved evidence, excluding answer
generation, at batch size 1 with resident model and index; neural methods
run on one H200 and BM25 on the host CPU.

\begin{table*}[t]
\centering
\caption{\textbf{Natural comparison-question retrieval and end-to-end RAG.}
Every retriever receives the same per-question distractor corpus and a total
budget of two sentences; all retrieved evidence goes to the same frozen
Qwen2.5-7B-Instruct reader. Accuracy in percent; time in ms per query, from
question to retrieved evidence, excluding answer generation. Shaded cells mark
the best non-oracle method in each column.}
\label{tab:natural-comparison-rag}
\vspace{3pt}
\small
\setlength{\tabcolsep}{2.4pt}
\renewcommand{\arraystretch}{1.05}
\resizebox{\textwidth}{!}{%
\begin{tabular}{lrrrrrrrrrr}
\toprule
& \multicolumn{5}{c}{\textbf{HotpotQA} ($n=1{,}296$)}
& \multicolumn{5}{c}{\textbf{2WikiMultiHopQA} ($n=2{,}933$)} \\
\cmidrule(lr){2-6}\cmidrule(lr){7-11}
Method & Sup.\,R@2$\uparrow$ & Both@2$\uparrow$ & EM$\uparrow$ & F1$\uparrow$ & ms$\downarrow$
       & Sup.\,R@2$\uparrow$ & Both@2$\uparrow$ & EM$\uparrow$ & F1$\uparrow$ & ms$\downarrow$ \\
\midrule
\multicolumn{11}{@{}l}{\textit{Shared baselines}} \\
BM25
  & 52.10 & 25.15 & 52.31 & 61.23 & \best{0.38}
  & 56.50 & 26.93 & 61.61 & 62.18 & \best{0.38} \\
Qwen2.5 pooling
  & 18.46 & 3.32 & 45.06 & 53.19 & 31.43
  & 14.31 & 1.94 & 38.56 & 43.20 & 31.44 \\
PromptReps (20/20), hybrid
  & 41.86 & 11.34 & 49.00 & 56.94 & 456.18
  & 44.85 & 13.16 & 55.74 & 56.33 & 456.94 \\
\midrule
\multicolumn{11}{@{}l}{\textit{Dream-7B-Instruct}} \\
Whole-query pooling
  & 32.91 & 11.11 & 47.61 & 56.07 & 29.33
  & 36.25 & 11.18 & 52.27 & 54.64 & 29.37 \\
DiffRetriever (4/16), dense
  & 40.33 & 13.97 & 51.00 & 59.94 & 32.29
  & 27.82 & 6.65 & 39.99 & 44.14 & 32.28 \\
DiffRetriever (4/16), hybrid
  & 44.86 & 16.28 & 50.77 & 59.60 & 35.85
  & 37.17 & 9.00 & 44.66 & 47.13 & 39.03 \\
\addlinespace[2pt]
Subject (L0)
  & \best{58.33} & \best{41.82} & 58.18 & 66.31 & 2.64
  & \best{59.34} & \best{37.37} & \best{65.60} & \best{66.26} & 2.64 \\
Subject (L27)
  & 39.17 & 20.76 & 55.56 & 63.67 & 31.95
  & 34.24 & 12.04 & 59.77 & 60.06 & 31.95 \\
\midrule
\multicolumn{11}{@{}l}{\textit{LLaDA-8B-Instruct}} \\
Whole-query pooling
  & 42.19 & 16.44 & 53.32 & 61.41 & 33.10
  & 35.90 & 10.13 & 50.90 & 53.02 & 33.08 \\
DiffRetriever (4/16), dense
  & 39.89 & 14.04 & 50.85 & 59.56 & 34.40
  & 33.16 & 6.65 & 43.23 & 45.70 & 23.47 \\
DiffRetriever (4/16), hybrid
  & 46.85 & 19.21 & 53.01 & 61.57 & 93.40
  & 43.51 & 13.60 & 52.88 & 54.16 & 93.38 \\
\addlinespace[2pt]
Subject (L0)
  & 58.01 & 40.82 & 57.56 & 65.52 & 2.62
  & 56.60 & 33.17 & 63.55 & 64.14 & 2.63 \\
Subject (L31)
  & 56.98 & 41.20 & \best{61.34} & \best{69.81} & 35.44
  & 44.04 & 20.39 & 62.60 & 62.77 & 35.48 \\
\midrule
Closed-book reader
  & -- & -- & 55.48 & 63.09 & --
  & -- & -- & 47.87 & 48.41 & -- \\
Oracle evidence
  & -- & -- & 62.73 & 71.77 & --
  & -- & -- & 82.07 & 82.14 & -- \\
\bottomrule
\end{tabular}}
\par\vspace{2pt}
\begin{minipage}{\textwidth}
\footnotesize
\textit{Protocol.} Subject rows use the entity-position readout; L0 reads
the last subject token of the plain input, L27/L31 the mean subject state
under the retrieval prompt, selected on validation. DiffRetriever uses four
query and sixteen passage representations; PromptReps twenty of each.
\textit{Timing.} Median per query at batch size 1 with resident model and
index; neural methods on one H200, BM25 on the host CPU.
\end{minipage}
\end{table*}

Dream L0 provides the strongest evidence retrieval on both datasets
(Table~\ref{tab:natural-comparison-rag}). Its Both@2 exceeds Dream
DiffRetriever hybrid by 25.54 points on HotpotQA (paired 95\% CI
$[22.30,28.78]$) and by 28.37 points on 2WikiMultiHopQA ($[26.49,30.24]$).
The improved evidence allocation raises answer F1 by 6.71 points
($[4.02,9.48]$) and 19.14 points ($[17.12,21.22]$), respectively. Relative to
BM25, Dream L0 improves Both@2 by 16.67 points on HotpotQA and 10.44 points
on 2Wiki, together with EM gains of 5.87 and 3.99 points. On HotpotQA, the
LLaDA L31 subject state gives the strongest final answers, reaching 61.34 EM
and 69.81 F1; on 2Wiki, Dream L0 is the strongest non-oracle method on all
four metrics. Entity-specific subject states thus allocate a fixed retrieval
budget across the two targets of a natural comparison question, and the
improvement carries through to the generated answer.

\subsection{Causal intervention and matching controls}
\label{app:retrieval-causal}

The interventions in this subsection use Dream-7B-Instruct and the earlier,
film-only version of the controlled benchmark ($3{,}745$ cards, $5{,}196$
queries in the same single, adjacent and crossed layouts).

\begin{figure}[htbp]
\centering
\safeincludegraphics[width=\linewidth]{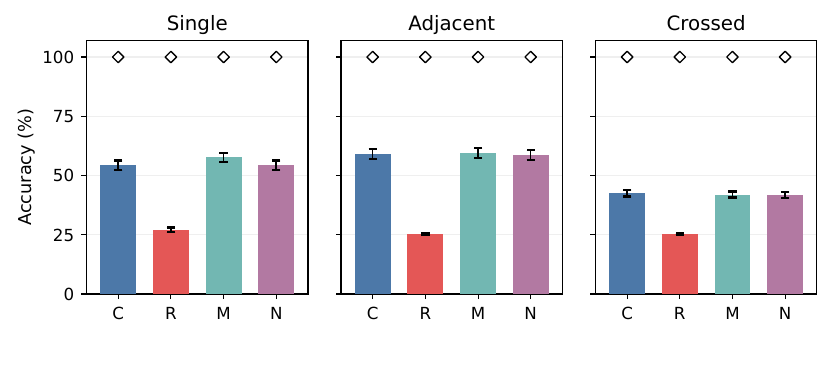}
\caption{\textbf{Relation messages support early evidence selection.}
Query-side layer-$0$ interventions on Dream with a fixed document index.
Bars: within-entity relation R@1 averaged over targets; diamonds: global
entity identification; error bars: entity/pair-clustered $95\%$ confidence
intervals. C: clean; R: relation messages removed; M: mass-matched random
messages (three-seed mean); N: norm-matched random messages.}
\label{fig:retrieval-causal}
\end{figure}

\paragraph{Summary.}
The dependence on relation fusion is causal. On Dream, we subtract the first
attention layer's relation-token value messages at the subject positions,
keeping all other attention weights and the document index fixed.
Within-entity relation R@1 falls to the $25\%$ four-choice chance level in
all three layouts ($54.3\!\to\!27.1$ single, $59.1\!\to\!25.3$ adjacent,
$42.5\!\to\!25.4$ crossed), while mass-matched and norm-matched random
messages leave it at the clean level (Fig.~\ref{fig:retrieval-causal};
Holm-adjusted $p\simeq0.0003$ in each layout), and entity identification
stays at $100\%$. The same loss appears at the L1 MLP key, the representation
that \S\ref{sec:mechanism} identified as the lookup key
(Table~\ref{tab:retrieval-interventions}). Replacing the removed message with a norm-matched
message of a different relation from the same entity redirects retrieval
toward that relation: donor-relation selection rises from $14.89\%$ to
$56.38\%$ on single queries and from $33.83\%$ to $57.57\%$ on adjacent
queries. Early access states thus carry the relation information that selects
evidence, and reading them is the counterpart of the write in
\S\ref{sec:method}: the layer-$0$ routing that an edit must cover is what a
retrieval vector taken from the state already contains.

\paragraph{Protocol.}
The causal experiment uses plain queries in Dream Instruct. We recompute
queries and all document states in float32, keep the document index fixed,
and intervene only on query-side L0 attention. For subject position $i$ and
relation-token set $R$, let
\begin{equation}
m_R(i)=W_O\operatorname{concat}_{h}
\left[\sum_{t\in R}\alpha^{h}_{it}v^{h}_t\right].
\end{equation}
We subtract $m_R(i)$ from the attention output at every token in each subject
span. $R$ includes the visible relation-word tokens from all target clauses
in the query. Other attention weights remain unchanged: there is no softmax
renormalization. We read the resulting L0 residual, L2 residual, and actual
input to the L1 MLP down-projection (the MLP key). These have dimensions
3,584, 3,584, and 18,944, respectively. The clean float32 global scores differ
from the original bfloat16 scores by at most 0.87 percentage points; every
causal contrast uses the same float32 computation.

The thirteen conditions comprise clean input; relation-message removal;
three native random-message removals; three exact-mass random-message
controls; a residual-norm-matched random control; clean relation add-back;
wrong-relation add-back; former/latter-message removal; and all
non-subject-template-message removal. Random token sets exclude subject,
relation, reference, and special tokens. For each seed, subject position,
and attention head, we sample 256 same-cardinality subsets and select the
closest attention-mass match without retrieval scores or correctness labels.

The native random subsets have mean relative mass errors of 32.39\%, 32.68\%,
and 32.51\%. The exact controls reweight each selected set to the relation
mass before output projection, achieving maximum effective relative errors
below $2.4\times10^{-7}$. Their maximum scaling factor is
$1.97\times10^7$: these are reweighted message interventions rather than
native-edge deletions. Independently, the norm control scales a random
projected message to $\|m_R(i)\|_2$, matching the residual perturbation size
with relative error below $1.67\times10^{-7}$. We report both control
families because they constrain different aspects of the intervention.

\begin{table}[t]
\centering\small
\begin{tabular}{lrrr}
\toprule
Query-side L0 intervention & Single & Adjacent & Crossed \\
\midrule
Clean & 54.31 & 59.11 & 42.53 \\
Remove relation messages & 27.07 & 25.26 & 25.35 \\
Native random removal, mean of three & 54.73 & 59.07 & 42.39 \\
Exact-mass random message, mean of three & 57.61 & 59.51 & 41.97 \\
Norm-matched random message & 54.37 & 58.56 & 41.81 \\
Clean relation add-back & 54.31 & 59.11 & 42.53 \\
Wrong-relation add-back & 10.29 & 35.39 & 48.34 \\
Remove former/latter messages & 54.31 & 59.11 & 41.96 \\
Remove non-subject template messages & 24.94 & 25.04 & 25.03 \\
\midrule
L2 residual: clean / relation removal
& 51.12 / 35.17 & 47.87 / 31.42 & 37.91 / 29.14 \\
L1 MLP key: clean / relation removal
& 41.84 / 30.23 & 41.78 / 28.36 & 35.56 / 27.94 \\
\bottomrule
\end{tabular}
\caption{\textbf{Causal readout results: within-entity relation R@1} (\%).
The first nine rows read the L0 residual. Random-seed rows are averaged
only for display; all thirteen conditions are stored individually.
Former/latter removal is a no-op in layouts without these tokens.}
\label{tab:retrieval-interventions}
\end{table}

\paragraph{Content transfer and reference binding.}
The donor uses the same entity and template with a different requested
relation. Single-target donors cycle to another relation; dual-target donors
swap the two relations while preserving both entities and the word bag.
We replace $m_R(i)$ by the donor's relation contribution, aligned to the same
subject tokens and scaled to its original norm. Wrong-relation selection
increases by 41.49 points on single queries (95\% CI $[38.56,44.43]$) and
23.74 points on adjacent queries ($[20.75,26.91]$). In crossed queries it
decreases by 5.45 points ($[-7.16,-3.79]$): clean states already favor the
donor relation (47.95\%) over the intended relation (42.53\%). After transfer,
correct-relation accuracy rises to 48.34\%. Relation availability and reference
binding are therefore empirically separable. Removing the messages from
former/latter tokens has a small effect on this L0 subject readout; directly
reading or patching those tokens is a separate localization experiment.

\paragraph{Inference and implementation checks.}
The prespecified primary contrast is the mean of the three exact-mass
controls minus relation removal in L0 relation R@1, separately for each
layout. Differences are 30.54, 34.25, and 16.62 points, with 95\% paired
cluster-bootstrap intervals $[28.49,32.53]$, $[32.17,36.39]$, and
$[15.30,17.88]$. We use 10,000 bootstrap and sign-flip draws; Holm correction
across the three primary tests gives $p\simeq0.00030$ for each. Norm-matched
contrasts have the same direction. Global entity identification remains
100\% for clean and relation removal; the paired interval $[0,0]$ lies inside
the fixed $\pm2$-point equivalence margin. This is preservation of the
measured identification endpoint at ceiling, rather than identity of the
full entity representation.

Clean-message add-back restores all saved readouts exactly, serving as an
algebraic implementation check. Reconstructed attention contributions match
native scaled-dot-product attention with maximum relative error
$6.08\times10^{-5}$. A six-query preflight verifies every non-subject L0
position is unchanged; the full run saves subject readouts only. The audit
checks all 77 entity-conditioned result rows, all 5,196 causal queries at
three readouts, donor alignment, finite features, validation selections,
metric aggregation, and code/data hashes. The follow-up protocol was frozen
before these evaluations and reuses the already-inspected main-table test
set. The resulting claim concerns frozen-model factual evidence selection
and its causal dependence on early relation messages; answer generation and
latency are reported in App.~\ref{app:retrieval-natural}.

\section{Limitations}
\label{app:limitations}

We list the limitations we consider most material, in rough order of how much
they constrain the claims of the paper.

\paragraph{Free-generation transfer is improved, not solved.}
The QAEdit results of \S\ref{sec:qaedit} show that Multi-View produces a clear
improvement in free generation after a statement-only edit: $44.6\%$ and
$32.5\%$ of previously wrong answers become at least partially correct. We want to be explicit that this is not the same as having solved
statement-to-question transfer. In absolute terms the weighted scores are
$0.342$ (Dream) and $0.250$ (LLaDA); fully correct answers are $23.8\%$ and
$17.6\%$; and $55.4\%$ and $67.5\%$ of edits are still answered wrongly. On a
paraphrased question (the \emph{Gen.} split) the numbers fall further. A user who
edits a fact by stating it and then asks about it in natural question form should
therefore still expect the edited model to be wrong more often than right. The
contribution here is a measurable step on a metric where AR editors have been
brittle, not a deployable level of reliability.

\paragraph{The edit specification is structured.}
Although we describe the input as a declarative statement, the current construction assumes identifiable subject, relation, and object spans. Benchmarks such as CounterFact and BAKE provide this structure naturally, whereas arbitrary user statements may contain multiple predicates, qualifiers, or non-contiguous relations. Extending statement-level editing to such inputs requires a mechanism for proposition decomposition and span discovery, which is outside the scope of the present editor.

\paragraph{Retrieval assumes identifiable entity spans and two-entity questions.}
Both settings provide the entity spans: matched paragraph titles in the
natural setting, templated spans in the controlled benchmark. Questions with
more than two entities, implicit or nested mentions, or spans that must be
detected first are untested, and the natural setting ranks within a
question's official distractor context rather than an open corpus.

\paragraph{Global single-entity search favours representative-token and lexical retrievers.}
On the controlled benchmark, DiffRetriever hybrid and BM25 reach $97.7\%$ and
$92.8\%$ single R@1 against $89.1\%$ for the best access-state readout
(Table~\ref{tab:retrieval-popqa}). The access state's advantage is in
allocating evidence across entities; there the deeper readouts help in global
search and cost a full forward pass, while the first-block readout is best
where the candidate set is a question's own context.

\end{document}